\documentclass[11pt]{article}

\usepackage[final]{acl}
\usepackage{times}
\usepackage{latexsym}
\usepackage{listings}
\usepackage{threeparttable}
\usepackage{tcolorbox}
\tcbuselibrary{breakable}
\usepackage{tabularx}  
\usepackage{siunitx}
\usepackage{enumitem}
\usepackage{titletoc}
\usepackage{soul}
\usepackage{hyperref}
\usepackage{soulutf8}
\usepackage[utf8]{inputenc} 
\usepackage{CJKutf8}
\usepackage[T1]{fontenc}    
\usepackage{hyperref}     
\usepackage{url}         
\usepackage{booktabs}
\usepackage{amssymb}
\usepackage{amsfonts}       
\usepackage{nicefrac}       
\usepackage{microtype}      
\usepackage{xcolor, colortbl}         
\usepackage{graphicx}
\usepackage{amsmath}
\usepackage{algorithm}
\usepackage{algorithmic}
\usepackage{multirow}
\usepackage{multicol}
\definecolor{citecolor}{HTML}{0071BC}
\definecolor{linkcolor}{HTML}{ED1C24}
\usepackage{tcolorbox}
\usepackage{pifont}
\usepackage{varwidth}
\usepackage{amsthm}  

\usepackage{booktabs}
\usepackage{makecell}
\usepackage{inconsolata}
\usepackage{graphicx}
\usepackage{booktabs}
\usepackage{array}
\usepackage{makecell}
\usepackage{soul}
\sethlcolor{green!15}

\usepackage{mathtools}
\usepackage{amsthm}
\theoremstyle{plain}
\newtheorem{desideratum}{Desideratum}

\newcommand{\xmark}{\ding{55}}%

\title{\textit{Human} or \textit{LLM} as Standardized Patients? \\ A Comparative Study in Medical Education }

\author{
\textbf{Bingquan Zhang}$^{1,2}$\thanks{Equal contribution.},
\textbf{Xiaoxiao Liu}$^{2}$\footnotemark[1],
\textbf{Yuchi Wang}$^{2}$,
\textbf{Lei Zhou}$^{3}$,
\textbf{Qianqian Xie}$^{1}$\thanks{Corresponding authors.},
\textbf{Benyou Wang}$^{2}$\footnotemark[2]\\
$^1$ School of Artificial Intelligence, Wuhan University\\
$^2$ The Chinese University of Hong Kong, Shenzhen\\
$^3$ Freedom AI
}

\begin{document}
\maketitle
\begin{abstract}
Standardized patients (SPs) are indispensable for clinical skills training but remain expensive and difficult to scale. Although large language model (LLM)-based virtual standardized patients (VSPs) have been proposed as an alternative, their behavior remains unstable and lacks rigorous comparison with human standardized patients. We propose EasyMED, a multi-agent VSP framework that separates case-grounded information disclosure from response generation to support stable, inquiry-conditioned patient behavior. We also introduce SPBench, a human-grounded benchmark with eight expert-defined criteria for interaction-level evaluation. Experiments show that EasyMED more closely matches human SP behavior than existing VSPs, particularly in case consistency and controlled disclosure. A four-week controlled study further demonstrates learning outcomes comparable to human SP training, with stronger early gains for novice learners and improved flexibility, psychological safety, and cost efficiency. The code is publicly available at \url{https://github.com/FreedomIntelligence/EasyMED}.

\end{abstract}

\section{Introduction}
\label{sec:introduction}


Clinical reasoning and doctor–patient communication are essential skills in medical education~\cite{cleland2022researching}. Their development relies on repeated, interactive practice in realistic clinical settings.
Standardized patients, trained actors who consistently portray predefined clinical cases, are widely regarded as the gold standard for teaching and assessing these skills, particularly in Objective Structured Clinical Examinations (OSCEs)~\cite{sayers2024database, ma2023standardized}. While SP-based training enables safe and authentic clinical encounters, human SP programs are costly, labor-intensive, and difficult to scale, which limits training frequency and accessibility~\cite{zendejas2013cost}. Consequently, large language model (LLM) based virtual standardized patients (VSP) have emerged as a promising scalable alternative~\cite{du2024llms, ye2025multimodal}, due to their strong dialogue capabilities and broad world knowledge.



Despite recent progress, it remains unclear whether LLM-based VSP can support clinical skills training at a level comparable to human standardized patients. This question is difficult to answer due to persistent gaps in system design and evaluation that are misaligned with real SP training practice.
Most VSP frameworks conflate inquiry interpretation with response generation, leading to premature information disclosure, cross-turn instability, and limited support for intent-aware instructional feedback~\cite{du2024llms, ye2025multimodal, sirdeshmukh2025multichallenge}. Existing evaluations are largely coarse-grained, relying on synthetic dialogues or outcome-level metrics rather than authentic human SP–doctor interactions~\cite{fan2023nphardeval, waisberg2024large}. Moreover, systematic long-term comparisons with human standardized patients under matched training conditions remain rare~\cite{liu2025interactive,bodonhelyi2025modeling}, leaving the educational effectiveness of LLM-based virtual patients insufficiently validated.

\begin{table*}[t]
\centering
\small
\renewcommand{\arraystretch}{1.2}

\resizebox{1.0\textwidth}{!}{
\begin{tabular}{l|ccc|cccc}%

\toprule
 \textbf{SP}
& \textbf{Patient Fidelity}
& \textbf{Interaction Coherence}
& \textbf{Pedagogical Awareness} 
& \textbf{Real-world User Study} \\
\midrule

Human SP
& \checkmark
& \checkmark
& \checkmark
& - \\
\midrule

SimPatient~\cite{steenstra2025scaffolding}
& \xmark
& \xmark
& \checkmark 
& \xmark \\

EvoPatient~\cite{du2024llms}
& \xmark
& \xmark
& \xmark
& \xmark \\

CureFun~\cite{li2024leveraging}
& \xmark
& \xmark
& \checkmark
& \xmark  \\
\midrule


\multirow{3}{*}{\textbf{EasyMED (Ours)}} 
& \checkmark 
& \checkmark 
& \checkmark 
& \checkmark    \\

& Patient  Agent & Auxiliary Agent 
& Evaluation Agent
&   \multirow{2}{*}{ (Sec.~\ref{sec:Experiment_protocol}) }   
\\

& intent-conditioned disclosure  & factorized patient simulation  
& trajectory-level feedback
&  
\\

\bottomrule
\end{tabular}
}

\caption{
Comparison of various standardized patient systems across three desiderata defined in Section~\ref{sec:background}, based  on what is explicitly reported in the original papers.
\textbf{SimPatient} evaluates realism primarily via qualitative user studies and adopts end-to-end patient response generation, while providing utterance-level reflective feedback.
\textbf{EvoPatient} emphasizes emergent realism through agent co-evolution without explicit control over information disclosure or learner-facing instructional feedback.
\textbf{CureFun} focuses on educational usefulness without controlled comparison to human standardized patients and performs end-to-end patient simulation, while offering post-session learning summaries. \textbf{EasyMed} embeds patient fidelity, interaction coherence, and pedagogical awareness directly into its architecture via intent-conditioned disclosure calibrated by SPBench  (Sec.~\ref{sec:spbench}), factorized patient simulation using Auxiliary Agent (Sec.~\ref{sec:patient agent}), and trajectory-level evaluation in Evaluation Agent  (Sec.~\ref{sec:evaluation agent}).
}
\label{tab:sp_comparison}
\vspace{-8pt}
\end{table*}

\noindent \textbf{Multi-agent VSP}
To address the limitations identified above, We propose EasyMED, a controllable multi-agent framework that models virtual SP training as a structured, interactive process. EasyMED decouples patient simulation, intent recognition, and evaluation into coordinated agents, enabling intent-conditioned information disclosure, stable cross-turn behavior, and checklist-based instructional feedback. This design directly supports patient fidelity, interaction coherence, and pedagogical awareness in virtual SP training.

\noindent \textbf{SP Benchmark}
To support reproducible and interaction-level evaluation, we further introduce SPBench, a benchmark constructed from authentic standardized patient-doctor dialogues spanning 14 medical specialties and eight expert-defined evaluation criteria~\cite{fan2023nphardeval, sirdeshmukh2025multichallenge}. 
Unlike existing benchmarks that rely on synthetic dialogues or outcome-level scores, SPBench uses human SP interaction trajectories as reference to quantitatively compare virtual and human standardized patient behavior. 
Using SPBench, we compare EasyMED and representative existing VSPs against human SP interaction trajectories, and find closer alignment with human SP behavior, particularly in controlled disclosure and cross-turn consistency.


\noindent \textbf{Real-world User Study}
To assess educational effectiveness in real training settings, we conduct a four-week controlled study comparing EasyMED with human standardized patient training under matched content and scoring rubrics, directly addressing whether LLM-based virtual patients can achieve training effectiveness comparable to human SPs.

Our \textbf{contributions} are threefold: (1) we propose \textbf{EasyMED}, a multi-agent virtual standardized patient framework that enables controllable, interpretable patient simulation aligned with clinical training workflows; 
(2) we propose \textbf{SPBench}, a human-grounded benchmark that enables quantitative, interaction-level comparison between virtual and human standardized patient behavior; 
(3) we present a controlled, longitudinal \textbf{user study} comparing LLM-based and human standardized patient training under matched conditions.

\begin{figure*}[h]
    \centering
    \small
    \includegraphics[width=0.9\linewidth]{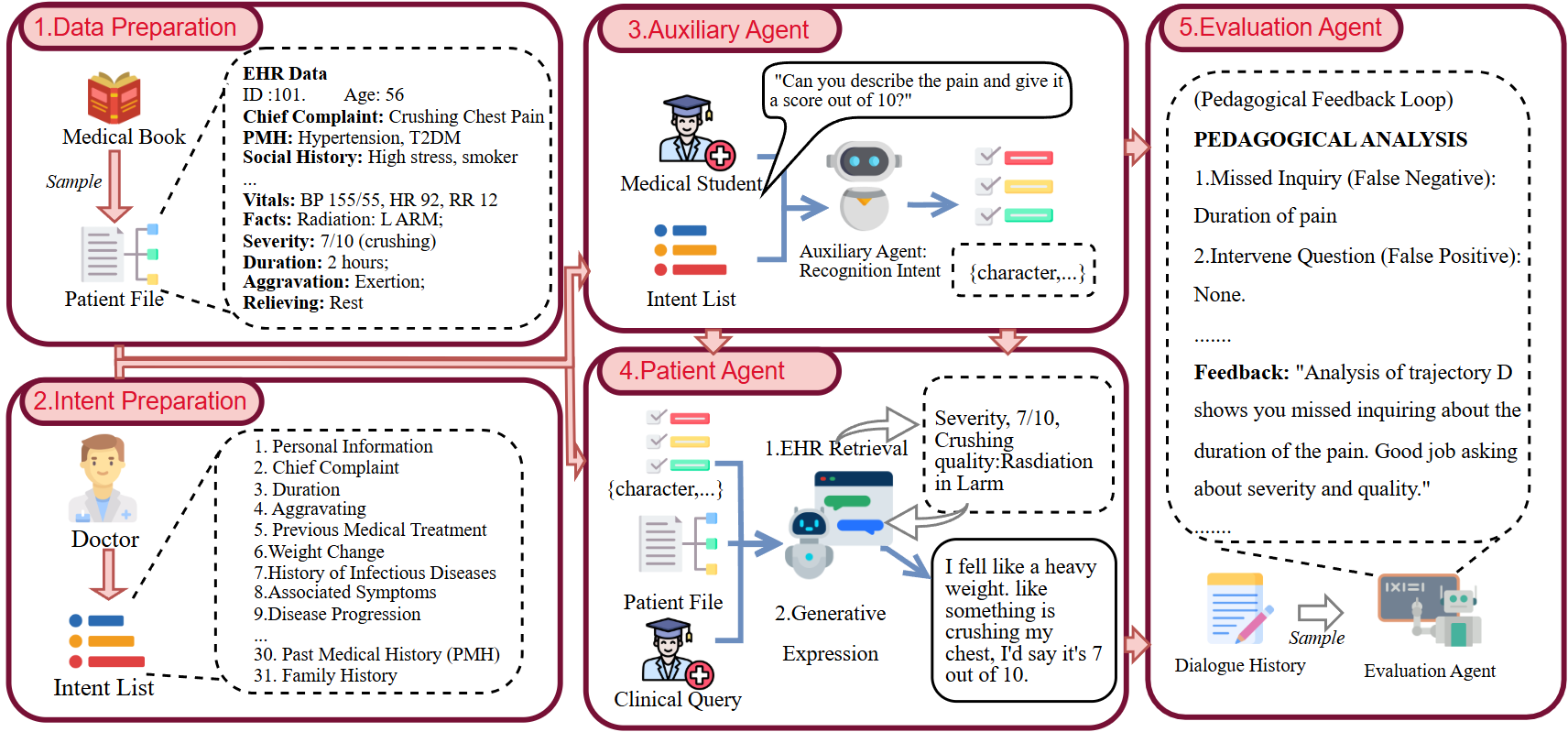}
    \caption{Overview of the multi-agent architecture of the virtual standardized patient system, consisting of a Patient Agent, an Intent Recognition Agent, and an Evaluation Agent.}
    \label{fig:agent_flow}
    \vspace{-10pt}
\end{figure*}

\section{Background}
\label{sec:background}




VSPs are increasingly adopted for clinical training due to their scalability and accessibility. However, their educational value hinges not only on their ability to simulate patient dialogue, but also on whether they can faithfully reproduce human standardized patient behavior (see Desideratum~\ref{desideratum:fidelity}), remain stable and controllable across extended interactions (see Desideratum~\ref{desideratum:coherence}), and actively support clinical learning feedback (see Desideratum~\ref{desideratum:feedback}). In practice, current VSPs often fall short in one or more of these aspects, limiting their reliability as training tools, see Table~\ref{tab:sp_comparison}. 

\begin{desideratum}\label{desideratum:fidelity}
\textbf{Patient Fidelity}.  VSPs should exhibit behaviors and response patterns comparable to those of human standardized patients. 
\end{desideratum}

Achieving such fidelity requires evaluation protocols that enable direct experimental comparison with human SPs, rather than relying solely on subjective user satisfaction. However, most prior studies on VSPs primarily adopt qualitative evaluations based on structured interviews and satisfaction surveys~\cite{steenstra2025scaffolding, du2024llms, li2024leveraging}. As a result, direct comparisons with human SPs remain rare, and the educational effectiveness of LLM-based VSPs relative to traditional human SP training is still unclear~\cite{simzine2025vsp}.

\begin{desideratum}\label{desideratum:coherence}
\textbf{Interaction Coherence}.
VSPs must maintain consistent clinical states and controlled information disclosure across multi-turn interactions.
\end{desideratum}

Interaction coherence requires separating what clinical information is revealed from how it is expressed, so as to prevent case drift and unintended information leakage across turns. However, most existing VSPs generate responses in an end-to-end manner without explicitly modeling this separation~\cite{steenstra2025scaffolding,li2024leveraging}. As a result, although responses may appear locally coherent, longer interactions often exhibit cross-turn instability and uncontrolled disclosure, undermining the reliability of VSPs for medical education.

\begin{desideratum}\label{desideratum:feedback}
\textbf{Pedagogical Awareness}.
VSP-based training systems should be aware of the learner’s educational objectives and interaction process, enabling fine-grained instructional feedback that supports clinical learning rather than merely simulating patient behavior.
\end{desideratum}

Pedagogical awareness requires VSPs to go beyond passive patient simulation and actively support learning by monitoring the learner’s inquiry process and identifying opportunities for guidance. Grounding feedback in the interaction trajectory allows such systems to highlight missing, redundant, or inappropriate clinical questions and support reflective learning. However, most existing VSPs function primarily as conversational agents and lack explicit representations of clinical intent or inquiry coverage, limiting their ability to provide meaningful instructional feedback~\cite{steenstra2025scaffolding,du2024llms}.




\section{EasyMED: A Multi-Agent VSP Framework}

\begin{table*}[tp]
\centering
\small
\setlength{\tabcolsep}{2pt} 
\begin{tabularx}{\textwidth}{@{}p{3.6cm}p{1cm}X@{}}
\toprule
\textbf{Criterion} & \textbf{Abbr.} & \textbf{Description} \\
\midrule
Query Comprehension & QC &
Accurate understanding of the physician’s question and its  intent without misinterpretation \\

Case Consistency & CC &
Faithfulness to the predefined patient case, without contradictions or unsupported facts. \\

Controlled Disclosure & CD &
Providing only requested information, avoiding unsolicited or premature disclosure. \\

Response Completeness & RC &
Fully addressing all aspects of the physician’s query without omitting essential case information. \\

Logical Coherence & LC &
Internal logical consistency of responses, ensuring symptoms and attributes remain coherent. \\

Language Naturalness & LN &
Use of natural, patient-like language while avoiding unnecessary medical jargon. \\

Conversational Consistency & CS &
Consistency of information across dialogue turns, avoiding self-contradictions. \\

Patient Demeanor & PD &
Maintaining an appropriate patient-like emotional tone, including cooperation and stability. \\
\bottomrule
\end{tabularx}
\caption{Definitions of the eight evaluation criteria used in SPBench.}
\label{tab:evaluation_dimensions}
\vspace{-6pt}
\end{table*}


\subsection{Workflow of EasyMED}

\paragraph{Philosophy of EasyMED}
Section~\ref{sec:background} identifies three desiderata for virtual standardized patients—patient fidelity, interaction coherence, and pedagogical awareness—that are difficult to achieve with end-to-end simulators. EasyMED treats these desiderata as explicit design constraints and maps them to concrete architectural choices: intent-conditioned response generation to preserve patient fidelity, decoupled case-grounded information access and surface realization to ensure interaction coherence, and trajectory-level retention for checklist-based educational feedback. This principled mapping naturally motivates a factorized, multi-agent design.

EasyMED implements a factorized workflow with two phases: \textit{consultation} and \textit{evaluation}, as shown in Figure~\ref{fig:agent_flow}. Phase 1 is realized by the Auxiliary and Patient Agents for intent recognition and patient simulation, while Phase 2 is conducted by the Evaluation Agent for trajectory-level assessment and feedback.

\noindent \textbf{Phase 1: Consultation } 
During consultation, the interaction unfolds as a multi-turn dialogue
\begin{equation}
\mathcal{D} = \{(q_1, r_1), \ldots, (q_T, r_T)\},
\end{equation}
where $q_t$ is the learner’s question and $r_t$ the patient response at turn $t$.
At each turn, EasyMED first infers a standardized clinical intent
\begin{equation}
i_t = A(q_t, H_{t-1}),
\end{equation}
and then generates a case-grounded response
\begin{equation}
r_t = P(i_t, q_t, H_{t-1} \mid E).
\end{equation}
By factorizing intent recognition and response generation, EasyMED enables inquiry-conditioned disclosure and stable multi-turn patient behavior.

\noindent \textbf{Phase 2: Evaluation } After the consultation ends, the Evaluation Agent reviews the full dialogue trajectory $\mathcal{D}$ and compares the recognized intents and elicited facts against expert-defined case checklists to produce structured feedback.

\subsection{Auxiliary Agent}
\label{sec:auxiliary agent}


The Auxiliary Agent addresses a core limitation of existing virtual standardized patients by explicitly modeling the learner’s clinical inquiry rather than relying on end-to-end text generation. It maps each learner question to a predefined clinical intent (e.g., Chief Complaint or Onset), abstracting away surface-level linguistic variation. This standardized intent representation serves as a control signal for downstream patient simulation, ensuring that responses are conditioned on inquiry type rather than phrasing, thereby enabling controlled information disclosure and stable patient behavior across multi-turn interactions.


\subsection{Patient Agent}
\label{sec:patient agent}




The Patient Agent simulates patient behavior while enforcing case fidelity and disclosure constraints. Given an inferred clinical intent, it first retrieves the corresponding fact from a structured electronic health record that defines the patient’s ground-truth case information, and then generates a natural language response conditioned on both the retrieved fact and a predefined patient persona (e.g., anxiety or hesitation). This separation between fact selection and surface realization enables controlled information disclosure while preserving natural conversational flow.




\subsection{Evaluation Agent}
\label{sec:evaluation agent}

The Evaluation Agent acts as a post-hoc pedagogical observer. Rather than intervening during conversation, it reviews the interaction history once the session ends. By comparing recognized intents and elicited facts against the standard case checklist, the agent generates structured feedback to highlight gaps like Missed Inquiries. This provides students with actionable feedback, addressing the evaluation limitations discussed in Section~\ref{sec:background}.

\section{SPBench: Benchmark for Virtual SP}



\subsection{The Philosophy of SPBench}

Existing medical benchmarks primarily assess static knowledge or aggregate outcomes and fail to capture interactive patient behavior (e.g., MMLU~\cite{hendrycks2021ethics}, MedQA~\cite{jin2021disease}), forcing VSP evaluation to rely on synthetic or interview-based data~\cite{fan2023nphardeval, waisberg2024large}. 
SPBench fills this gap by grounding evaluation in authentic human SP–doctor dialogue trajectories and adopting a standardized protocol with eight clinically motivated dimensions that assess both turn-level response quality and session-level behavioral consistency.

\subsection{Data Curation}


\textbf{Data Structure} SPBench contains two main parts: (1) Patient Profile: These describe the patient's main complaint, symptoms, history, and background; and (2) Authentic Dialogue Records: turn-level transcripts showing how trained human SP respond to clinical questions in real instructional settings. We use these records as a standard. They allow us to compare the AI's performance against a human actor handling the same case.


\begin{figure}[h]
    \centering
    \includegraphics[width=1\linewidth]{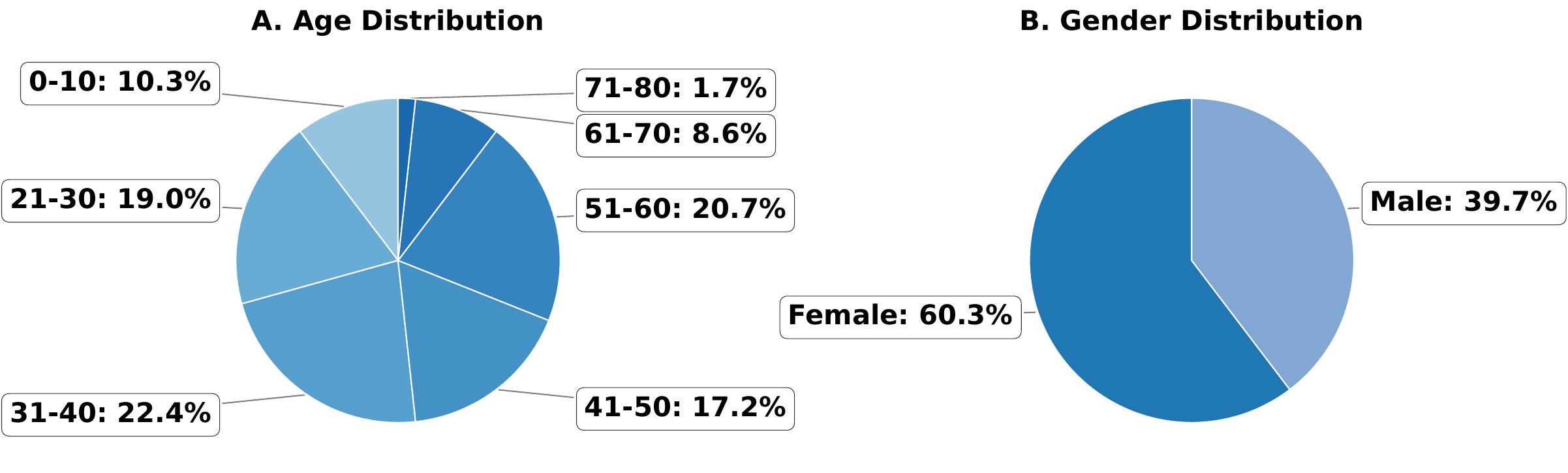}
    \caption{Demographic distribution of cases in the SPBench dataset. The left panel shows the age distribution, and the right panel shows the gender distribution.}
    \label{fig:SPBench2}
\end{figure}

\begin{table*}[t]
\centering
\small

\setlength{\tabcolsep}{6pt}
\begin{tabular}{lccccccccc}
\toprule
\textbf{System}  &
\textbf{QC} & \textbf{CC} & \textbf{CD} & \textbf{RC} &
\textbf{LC} & \textbf{LN} & \textbf{CS} & \textbf{PD} & \textbf{Overall}  \\
\midrule
\rowcolor{gray!10} \multicolumn{10}{c}{\textit{LLMs}} \\
\midrule
\textbf{Qwen3-8B}              & 77.76 & 78.64 & 84.26 & 85.74 & 83.37 & 72.14 & 85.74 & 81.31 & 81.12 \\
\textbf{Qwen3-32B}             & 88.37 & 88.37 & 89.31 & 89.31 & 88.99 & 85.89 & 89.91 & 89.61 & 88.87 \\
\textbf{DeepSeek-R1}           & 91.06 & 93.08 & 93.42 & 97.10 & \textbf{96.10} & 85.38 & 93.69 & 95.05 & 93.11 \\
\textbf{GPT-4o}                & 75.87 & 89.24 & 88.89 & \textbf{98.11} & 91.31 & 91.66 & 90.62 & 90.62 & 89.54 \\
\textbf{Gemini 2.5 Pro}       & \textbf{94.72} & \textbf{95.05} & \textbf{96.04} & 95.69 & 94.39 & \textbf{93.21} & \textbf{96.04} & \textbf{95.27} & \textbf{95.04} \\
\midrule

\rowcolor{gray!10} \multicolumn{10}{c}{\textit{Prompt Strategy + Agent Framework} }\\
\midrule
\textbf{Gemini 2.5 Pro + CoT}   & 96.61 & 94.98 & 96.63 & 95.33 & 93.99 & 95.59 & 97.61 & 95.61 & 95.77 \\

\textbf{Human SP} (reference)  & 95.71 & 96.47 & \textbf{98.53} & \textbf{97.85} & \textbf{98.18} & 95.10 & \textbf{98.56} & \textbf{98.22} & \textbf{97.33}\\
\textbf{EvoPatient}            & 91.87 & 96.29 & 94.59 & 95.77 & 92.67 & 90.49 & 92.32 & 92.74 & 93.33 \\
\textbf{EasyMED} (ours) & \textbf{97.17} & \textbf{97.23} & 98.18 & 97.11 & 95.03 & \textbf{97.48} & 97.98 & 95.73 & \textbf{96.98}\\
\bottomrule
\end{tabular}
\caption{
Overall and per-dimension performance evaluation on SPBench. Human SP serves as the gold standard reference. The best-performing LLM-based system is highlighted in bold.
}
\vspace{-10pt}
\label{tab:overall_spbench}
\end{table*}

\noindent \textbf{Case Source and Scope} SPBench is derived from two widely used training books for Standardized Patients—the \textit{Manual for Writing Standardized Patient Cases}\footnote{https://www.pmph.com/} and the \textit{A Practical Tutorial for Standardized Patients}\footnote{https://www.pumpedu.com/home-shop/7125.html}. From these resources, we collected 3,208 question–answer pairs used in real doctor–patient interactions. We cleaned and refined this data into 58 separate patient cases. Each case includes a profile and its dialogue records. The dataset is intended exclusively for academic research and evaluation purposes.

\begin{figure}[h]
\vspace{-10pt}
    \centering
    \small
    \includegraphics[width=0.9\linewidth]{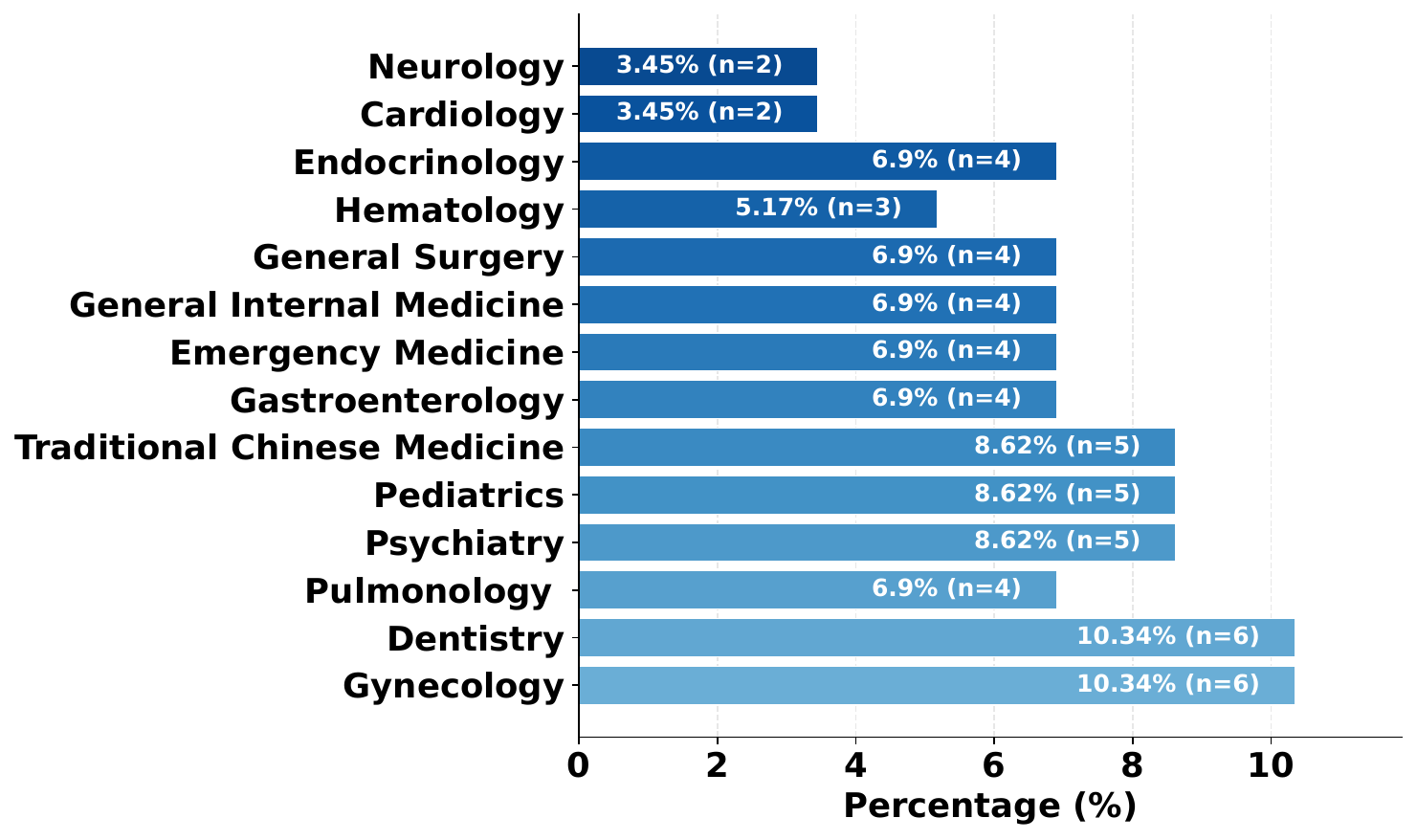}
    \caption{Distribution of clinical cases in the SPBench dataset by medical department.}
    \label{fig:SPBench_data_ditribute}
    \vspace{-15pt}
\end{figure}

Two clinical experts (Appendix~\ref{sec:partner}) checked each case to ensure anonymity, realism, and pedagogical usefulness. As shown in Figures~\ref{fig:SPBench_data_ditribute} and~\ref{fig:SPBench2}, SPBench covers 14 medical fields. 


\noindent \textbf{Quality Control} To ensure accuracy and reliability, we scanned the books using Optical Character Recognition (OCR). Then, three senior medical students manually verified the extracted text and corrected typographical and punctuation errors. This step ensures the data faithfully matches the original books, which is necessary for a reliable benchmark.


\subsection{Evaluation Protocol}

\noindent \textbf{Evaluation Metric}
SPBench evaluates VSP by assessing both turn-level response quality and session-level interactional behavior. Unlike existing benchmarks that rely on static knowledge tests or single aggregate scores, our evaluation decomposes patient performance into multiple clinically interpretable dimensions.

Specifically, we define eight evaluation criteria (Table~\ref{tab:evaluation_dimensions}) in collaboration with three clinical experts (see Appendix~\ref{sec:partner}). These criteria capture complementary aspects of patient simulation, including accurate understanding of clinical inquiries, adherence to the predefined case, controlled disclosure of information, and consistency across dialogue turns. Each criterion is independently rated on a 5-point Likert scale, and scores are linearly rescaled to a 100-point scale for reporting.




\noindent\textbf{Input Standardization}
To ensure fair and reproducible evaluation, SPBench uses authentic questions extracted from real-world doctor--patient dialogues. For each clinical case, we isolate the sequence of questions asked by the human physician and present this exact sequence to each model. This design eliminates variability introduced by different prompting styles and ensures that all virtual standardized patients are evaluated under identical input conditions.


\section{Evaluation on SPBench}
\label{sec:spbench}

This section analyzes divergences between virtual and human standardized patients in multi-turn interactions and assesses how EasyMED reduces these gaps relative to human SP behavior.

\noindent \textbf{Overall Performance}~
Table~\ref{tab:overall_spbench} summarizes overall and per-dimension performance on SPBench. Human standardized patients achieve the highest reference score (97.33), reflecting stable case portrayal and appropriate information disclosure across turns.
EasyMED closely matches human performance (96.98), with strongest gains on interaction-critical dimensions (CC, CD, CS, PD) that directly align with standardized patient requirements. By contrast, although several frontier LLMs perform well on LN and RC, they show larger variance on CC and CD, indicating unstable case grounding and inconsistent inquiry-conditioned disclosure.



\noindent \textbf{Sensitivity on Prompting Strategies}~
To disentangle the sources of performance differences, we first examine LLM baselines under a unified prompting scheme. Although large models such as Gemini 2.5 Pro achieve relatively balanced scores, consistent weaknesses remain in controlled disclosure and cross-turn stability.
We further evaluate common prompting strategies using a fixed backbone (Gemini 2.5 Pro). CoT prompting improves reasoning transparency and modestly increases CC and RC. However, it also amplifies verbosity and unsolicited explanation, leading to reduced CD scores. Overall, prompting mainly influences response articulation rather than information control, and is insufficient to enforce standardized patient behavior in multi-turn interactions.



\noindent \textbf{Ablation Study on Auxiliary Agent}~
We next examine the effect of the Auxiliary Agent in EasyMED, which decouples intent recognition from response generation. As shown in Table~\ref{tab:overall_spbench}, this component yields the largest gains on CC, CD, CS, and PD. 
By mapping learner queries to standardized clinical intents, the Auxiliary Agent provides an explicit control signal that constrains information access, mirroring how human standardized patients condition responses on inquiry type rather than full case narratives. These gains exceed those from prompt engineering alone, highlighting the dominant role of architectural control in stabilizing multi-turn patient behavior.


\begin{figure*}[t]
    \centering
    \small
    \includegraphics[width=0.75\linewidth]{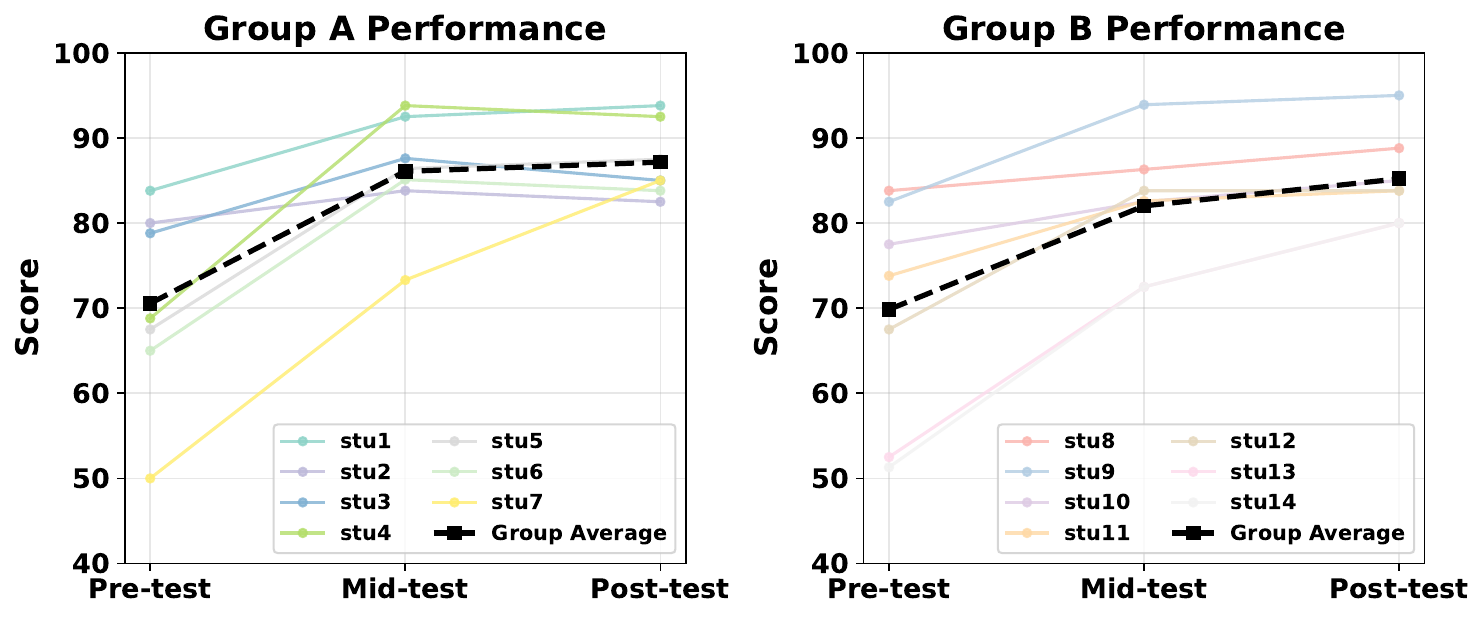}
    \caption{Learning trajectories of individual students and group averages for Group A (left) and Group B (right) across three assessment points. Each colored line tracks an individual student's performance, while the bold dashed line represents the group's average score. }
    \label{fig:student_performance}
\end{figure*}

\section{Real-world Evaluation in Medical Education}
\label{sec:Experiment_protocol}

While SPBench evaluates interaction-level fidelity, it does not directly capture educational effectiveness. We therefore assess EasyMED in a real training setting through a controlled user study, comparing it with human standardized patient training in terms of learning outcomes, learner experience, and practical feasibility.

\begin{table}[h]

\centering
\small
\resizebox{\linewidth}{!}{
\begin{tabular}{llll}
\toprule
\textbf{Study Period} & \textbf{Timeline} & \textbf{Group A} & \textbf{Group B} \\
\midrule
Baseline & Week 0 & \multicolumn{2}{c}{Pre-Test} \\
\midrule
\textbf{Period 1} & Weeks 1-2 & EasyMED Training & Human SP Training \\
\cmidrule{2-4}
& End of Week 2 & \multicolumn{2}{c}{Mid-Test} \\
\midrule
\textbf{Period 2} & Weeks 3-4 & Human SP Training & EasyMED Training \\
\cmidrule{2-4}
& End of Week 4 & \multicolumn{2}{c}{Final Test \& Questionnaire} \\
\bottomrule
\end{tabular}
}
\caption{Experimental design of the four-week study, showing the sequence of training interventions and assessments for Group A and Group B.}
\label{tab:crossover_design}
\vspace{-15pt}
\end{table}
\subsection{Experimental Design}

We adopted a randomized crossover design to enable within-subject comparison while controlling for baseline differences. Each participant experienced both EasyMED and human SP training in different phases, which allows analysis of overall learning gains as well as phase-specific effects attributable to each modality (Table~\ref{tab:crossover_design}).




\begin{table*}[tp]
\centering
\small
\resizebox{0.8\textwidth}{!}{
\begin{tabular}{l ccc ccc}
\toprule
\multirow{2}{*}{\textbf{Group Sequence}} & \multicolumn{3}{c}{\textbf{Mean Score (±SD) at Test Point}} & \multicolumn{3}{c}{\textbf{Mean Score Gain (±SD) by Phase}} \\
\cmidrule(lr){2-4} \cmidrule(lr){5-7}
& \textbf{Pre-test} & \textbf{Mid-test} & \textbf{Post-test} & \textbf{Phase 1 (Wks 1-2)} & \textbf{Phase 2 (Wks 3-4)} & \textbf{Total Gain} \\
\midrule
\textbf{Group A (AI$\rightarrow$SP)} & 70.56 & 86.07 & 87.44 & \textbf{+15.51 (AI)} & \textbf{+1.37 (SP)} & +16.89 \\
(N=7) & (±11.45) & (±6.88) & (±4.53) & (±7.82) & (±3.65) & (±7.45) \\
\midrule
\textbf{Group B (SP$\rightarrow$AI)} & 69.84 & 82.01 & 85.20 & \textbf{+12.17 (SP)} & \textbf{+3.19 (AI)} & +15.36 \\
(N=7) & (±13.04) & (±7.42) & (±4.93) & (±7.45) & (±3.25) & (±8.36) \\
\bottomrule
\end{tabular}
}
\caption{The table presents mean scores at each test point and the corresponding mean score gains during each training phase. Participants were in either Group A or Group B. All values are mean $\pm$ standard deviation.}
\label{tab:crossover_performance}
\vspace{-10pt}
\end{table*}

\noindent \textbf{Participants}
We recruited 20 medical undergraduate students in their fourth or fifth year. All participants completed a pre-test and received a 10-minute introduction to EasyMED. Students with scheduling conflicts or anomalous test scores were excluded (Appendix~\ref{sec:participant_exclusion}), yielding a final cohort of 14 students (7 men, 7 women; age range 21--24, mean 23). All participants had completed core clinical coursework but had not yet taken the National Medical Licensing Examination.

Participants were ranked by pre-test scores and assigned to groups using an alternating allocation scheme. Three experienced professionals served as human standardized patients. All participants were compensated on an hourly basis in accordance with institutional ethical guidelines.

\subsection{Results and Analysis}

\subsubsection{Evaluating Overall Improvement via OSCE}
\label{sec:osce}

\begin{figure}[h!]
    \centering
    \small
    \includegraphics[width=0.75\linewidth]{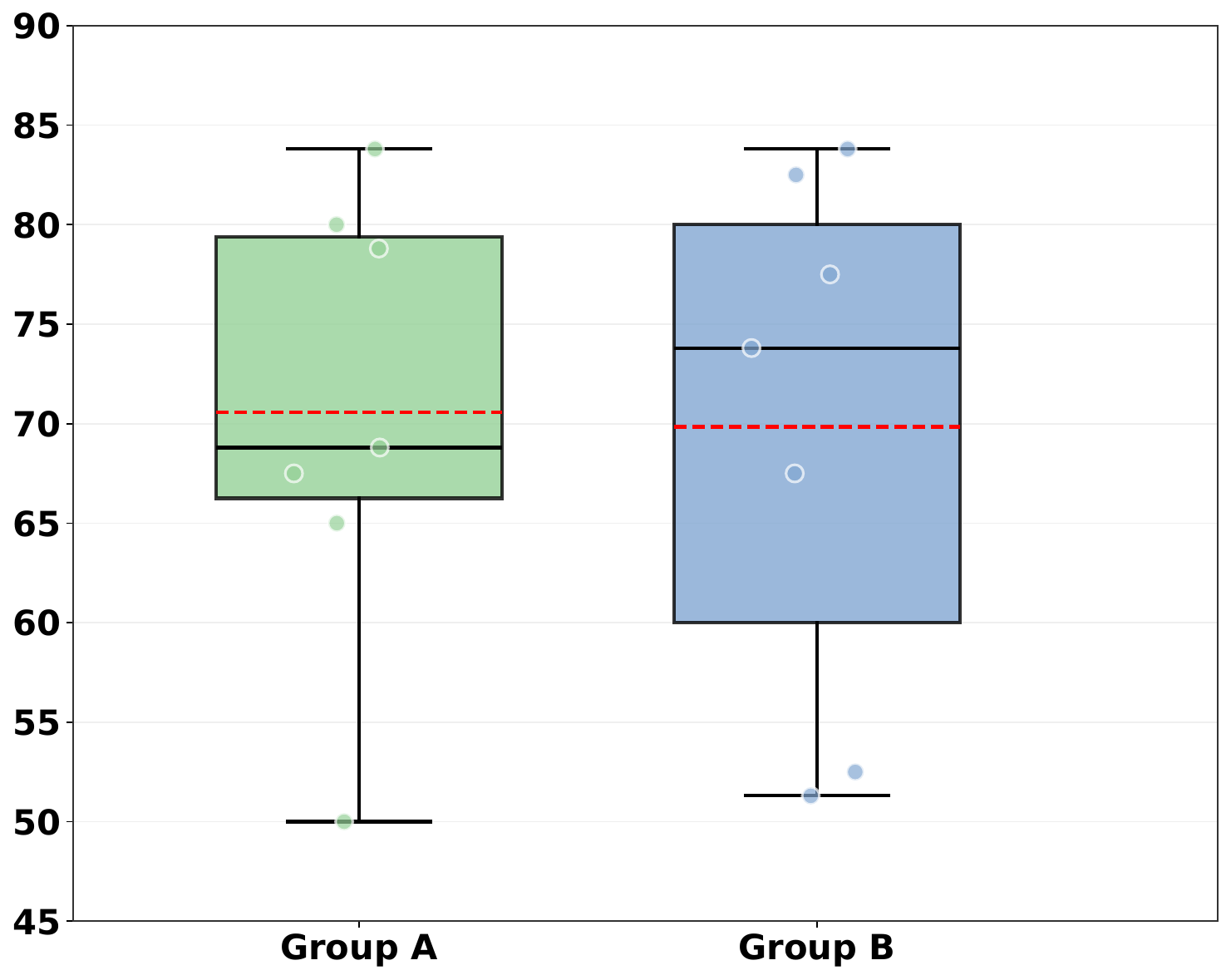}
    \caption{Boxplots of baseline OSCE scores for Group A and Group B prior to the intervention. Each point represents an individual participant.}
    \label{fig:inital score}
    \vspace{-5pt}
\end{figure}

Objective Structured Clinical Examination (OSCE) is a standardized, station-based clinical skills assessment.
We first confirm that the two groups were comparable prior to the intervention. As shown in Figure~\ref{fig:inital score}, baseline OSCE score distributions did not differ significantly between Group~A (mean = 70.56) and Group~B (mean = 69.84; $t(12)=0.16$, $p=0.88$), indicating similar starting levels.

\noindent \textbf{Overall~Performance}  Across the four-week study, both groups demonstrate substantial and comparable improvements in OSCE scores. As summarized in Table~\ref{tab:crossover_performance}, Group~A improved by 16.89 points on average, while Group~B improved by 15.36 points. Figure~\ref{fig:student_performance} shows consistent upward trends across individual learners in both groups. 

\hl{\textit{\textbf{Finding 1}: These results indicate that EasyMED supports clinical skill acquisition at a level comparable to human standardized patient training.}}




\begin{figure*}[ht]
    \centering
    \includegraphics[width=0.85\linewidth]{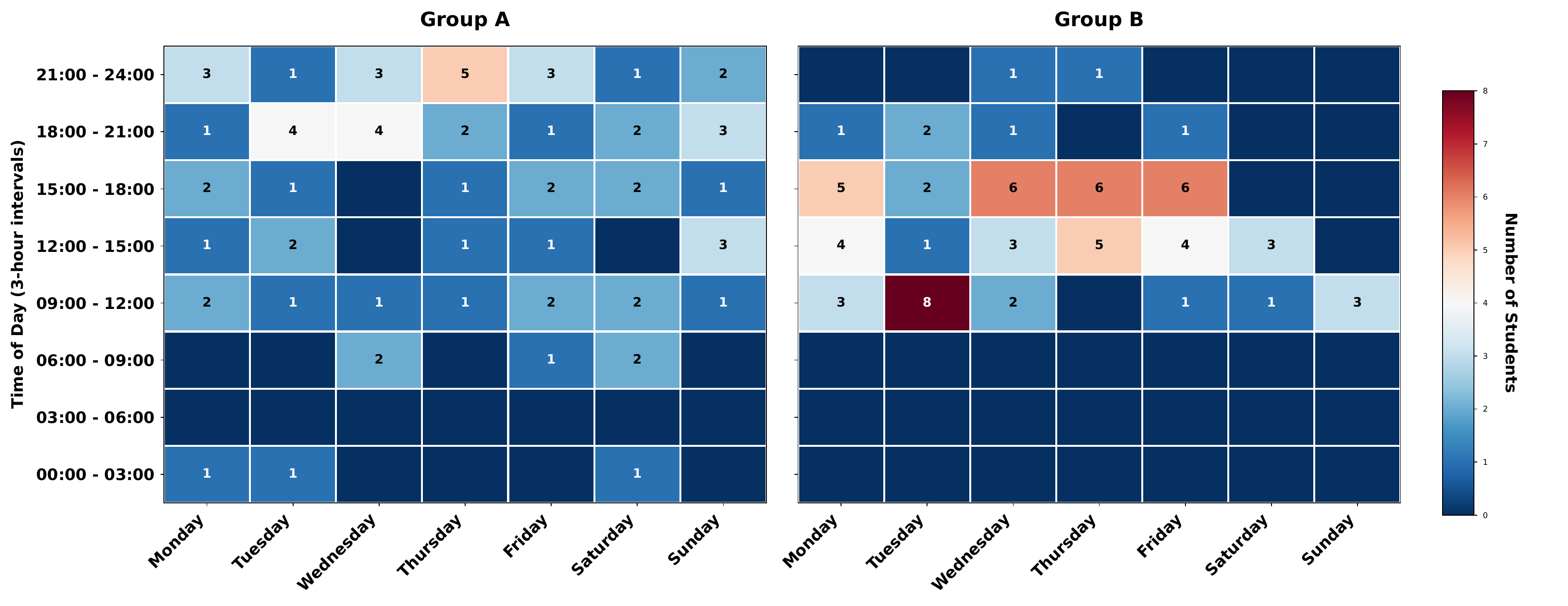}
    \caption{Comparative heatmap of the weekly practice time distribution for the \textit{EasyMED} and Human SP groups. 
    The left panel shows the \textit{EasyMED} group, and the right panel shows the Human SP group. 
    In both heatmaps, the x-axis represents the day of the week, and the y-axis represents the time of day. The color intensity and the white number in each cell indicate the number of students who practiced during that time slot. 
    }
    \label{fig:heatmap}
    \vspace{-10pt}
\end{figure*}

\noindent \textbf{Phase-wise Effects} Most learning gains occurred during the initial training phase for both modalities. During Phase~1 (Weeks~1--2), Group~A gained 15.51 points using EasyMED, while Group~B gained 12.17 points using human SPs. In Phase~2 (Weeks~3--4), when groups switched modalities, additional gains were observed but at a slower rate, suggesting diminishing returns commonly seen in short-term intensive training.

\begin{figure}[h!]
    \centering
    \includegraphics[width=0.75\linewidth]{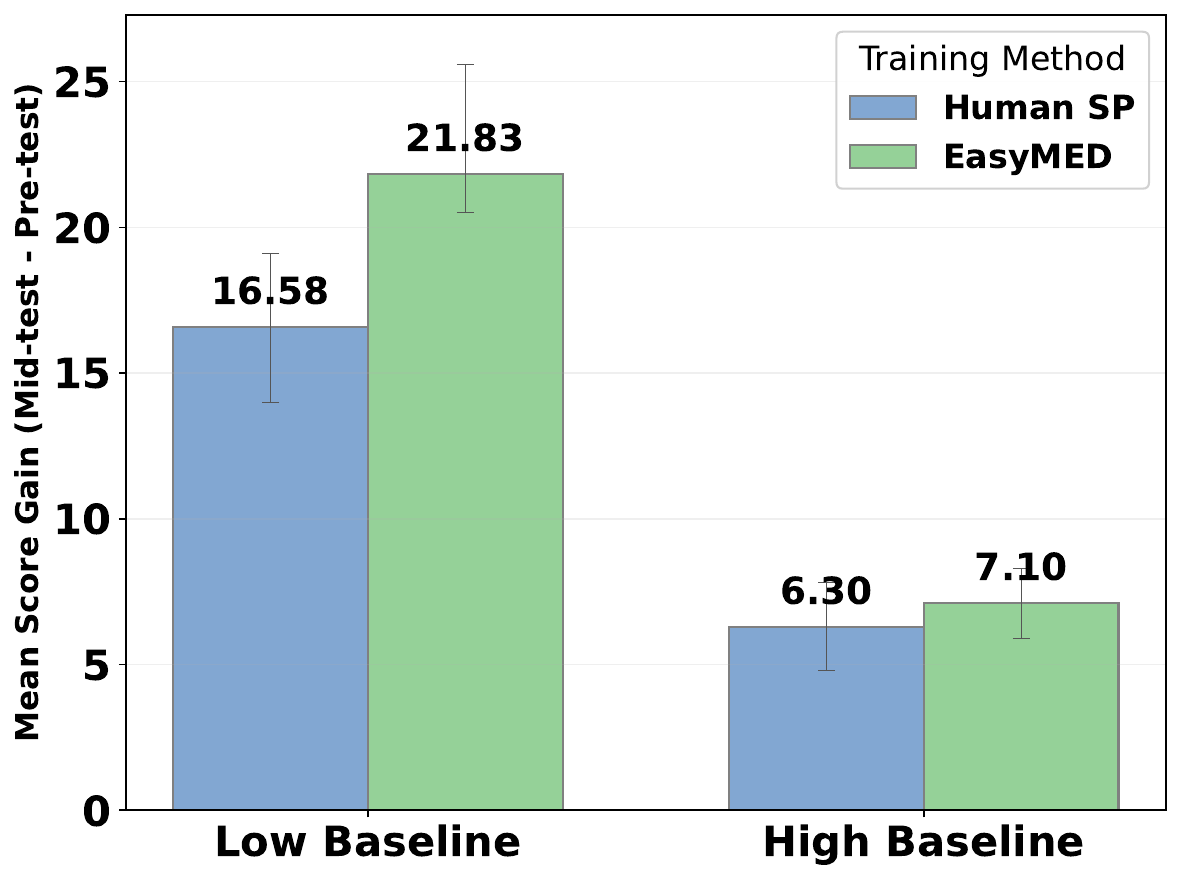}
    \caption{Comparison of mean score gains in Phase 1 for the Human SP and \textit{EasyMED} training methods. The participants are stratified into low- and high-performing groups based on their pre-test scores. Error bars indicate the standard error of the mean.}
    \label{fig:baseline_gain_comparison_phase1}
    \vspace{-15pt}
\end{figure}

\noindent \textbf{Improvement by Skill Level} To examine individual differences, we stratify participants into high-baseline (top three) and low-baseline (bottom four)  groups based on their pre-test OSCE scores. As shown in Figure~\ref{fig:baseline_gain_comparison_phase1}, low-baseline using EasyMED gained an average of 21.83 points, compared to 16.58 points with human SP. High-baseline improved less (7.10 vs. 6.30). 

\hl{\textit{\textbf{Finding 2}: This indicates that EasyMED is particularly effective for novice learners during the early stage of training (i.e., the first two weeks).}}




\subsubsection{Behavioral Analysis via Survey and Logs}

To further contextualize the learning outcomes reported in Section~\ref{sec:osce}, we analyzed students’ subjective questionnaire responses (see Appendix~\ref{sec:question}) together with interaction logs collected during training.

\noindent \textbf{Perceived Authenticity} Students reported high perceived realism when interacting with EasyMED. On a five-point Likert scale, the simulated patient dialogue achieved a mean authenticity score of 4.6 (Table~\ref{tab:interaction_comparison}), indicating that the interaction was generally regarded as natural and clinically plausible.

Ratings for learning helpfulness were comparable to those of human SP training, suggesting that EasyMED is perceived not merely as a convenient substitute, but as a viable modality for practicing history-taking and clinical reasoning.


\noindent \textbf{Peer Pressure} Survey results indicate substantially lower learning anxiety during EasyMED sessions than during human SP interactions (mean anxiety score 0.5 vs.~3.2, $p < .01$). Students reported feeling less concerned about making mistakes and more willing to ask exploratory or repeated questions. This low-pressure environment may facilitate risk-free exploration, particularly for learners at an early stage of training.


\noindent \textbf{Behavioral Evidence} System logs provide objective evidence that complements these subjective reports. As shown in Figure~\ref{fig:heatmap}, EasyMED practice sessions were distributed across a wide range of times, including evenings and weekends, whereas human SP sessions were largely confined to weekday working hours. In addition, EasyMED sessions involved more interaction on average, with a higher number of dialogue turns (54 vs. 47) and longer session durations (28:49 vs. 15:17) than human SP sessions (Table ~\ref{tab:interaction_comparison}). Although text-based interaction may partially account for longer durations, the increased number of turns suggests more iterative questioning and sustained engagement.

\begin{table}[h]
\centering
\small
\begin{tabular}{lll}
\toprule
\textbf{Metric} & \textbf{EasyMED} & \textbf{Human SP} \\
\midrule
\multicolumn{3}{l}{\textit{Student Engagement}} \\
Authenticity & 4.6 & -- \\
Helpfulness & 4.5 & 4.7 \\
Learning Anxiety Score\footnotemark & 0.5 & 3.2 \\
Average Dialogue Turns & 54 & 47 \\
Average Interaction Duration & 28m 49s & 15m 17s \\
\midrule
\multicolumn{3}{l}{\textit{Cost-Effectiveness}} \\
Per-Session Cost & \$0.725 & \$52.95 \\
\toprule

\end{tabular}

\caption{Comparison of student engagement and cost-effectiveness metrics between \textit{EasyMED} and human SP.}
\label{tab:interaction_comparison}
\vspace{-10pt}
\end{table}
\footnotetext{Anxiety was rated on a scale where lower scores are better. The difference is statistically significant ($p < .01$).}

\noindent \textbf{Cost-Effectiveness:} For cost estimation, EasyMED session costs are computed from the total token usage of a complete training interaction, whereas human SP costs are estimated by converting standard hourly compensation into a per-session cost. This results in an approximately 73-fold cost reduction compared to traditional SP training.

\hl{\textit{\textbf{Finding 3:} EasyMED provides a realistic, low-pressure, and accessible training environment that supports sustained and exploratory practice at a fraction of the cost of human SP training.}}


\section{Conclusion}

This study examines whether large language models can function as standardized patients for clinical skills training. We propose EasyMED, a multi-agent framework for stable, inquiry-conditioned patient simulation, and introduce SPBench, a human-grounded benchmark built from standardized patient–student dialogues. In a four-week controlled study, EasyMED achieves learning outcomes comparable to human SP training, with stronger early gains for novice learners, greater flexibility, and substantially lower cost. These results suggest that LLM-based multi-agent VSPs are a practical and scalable complement to traditional SP programs.

\section*{Limitations}

Our study has several limitations. It was conducted at a single institution with a relatively small and homogeneous cohort, so broader validation across different settings and learner populations is needed. In addition, EasyMED currently supports only text-based interactions without non-verbal or multimodal cues, which are important for authentic clinical communication. Finally, although our automated scoring showed strong correlation with expert ratings, it may still overlook subtle aspects of dialogue quality and learner behavior. Future work will include larger multi-site and longitudinal studies, integration of multimodal interaction channels, and refinement of evaluation metrics to better capture nuanced performance.

\section*{Ethical Statement}
The study was approved by the  institute on August 30, 2025. All annotators were fairly compensated, adhering to the standard hourly wage practices of their respective states.

\bibliography{custom}

@article{almansoori2025self,
  title={Self-Evolving Multi-Agent Simulations for Realistic Clinical Interactions},
  author={Almansoori, Mohammad and Kumar, Komal and Cholakkal, Hisham},
  journal={arXiv preprint arXiv:2503.22678},
  year={2025}
}

@article{berman2016role,
  title={The role for virtual patients in the future of medical education},
  author={Berman, Norman B and Durning, Steven J and Fischer, Martin R and Huwendiek, Soren and Triola, Marc M},
  journal={Academic medicine},
  volume={91},
  number={9},
  pages={1217--1222},
  year={2016},
  publisher={LWW}
}

@article{bodonhelyi2025modeling,
  title={Modeling Challenging Patient Interactions: LLMs for Medical Communication Training},
  author={Bodonhelyi, Anna and Stegemann-Philipps, Christian and Sonanini, Alessandra and Herschbach, Lea and Szep, Marton and Herrmann-Werner, Anne and Festl-Wietek, Teresa and Kasneci, Enkelejda and Holderried, Friederike},
  journal={arXiv preprint arXiv:2503.22250},
  year={2025}
}

@book{cleland2022researching,
  title={Researching medical education},
  author={Cleland, Jennifer and Durning, Steven J},
  year={2022},
  publisher={John Wiley \& Sons}
}

@article{cook2009virtual,
  title={Virtual patients: a critical literature review and proposed next steps},
  author={Cook, David A and Triola, Marc M},
  journal={Medical education},
  volume={43},
  number={4},
  pages={303--311},
  year={2009},
  publisher={Wiley Online Library}
}

@article{chen2020analysis,
  title={Analysis of learning behavior in an automated programming assessment environment: A code quality perspective},
  author={Chen, Hsi-Min and Nguyen, Bao-An and Yan, Yi-Xiang and Dow, Chyi-Ren},
  journal={IEEE access},
  volume={8},
  pages={167341--167354},
  year={2020},
  publisher={IEEE}
}

@article{du2024llms,
  title={LLMs Can Simulate Standardized Patients via Agent Coevolution},
  author={Du, Zhuoyun and Zheng, Lujie and Hu, Renjun and Xu, Yuyang and Li, Xiawei and Sun, Ying and Chen, Wei and Wu, Jian and Cai, Haolei and Ying, Haohao},
  journal={arXiv preprint arXiv:2412.11716},
  year={2024}
}

@article{fan2023nphardeval,
  title={Nphardeval: Dynamic benchmark on reasoning ability of large language models via complexity classes},
  author={Fan, Lizhou and Hua, Wenyue and Li, Lingyao and Ling, Haoyang and Zhang, Yongfeng},
  journal={arXiv preprint arXiv:2312.14890},
  year={2023}
}

@article{gai2024achieving,
  title={Achieving higher factual accuracy in llama llm with weighted distribution of retrieval-augmented generation},
  author={Gai, Zhenhua and Tong, Lianxin and Ge, Quan},
  year={2024}
}

@inproceedings{gan2023large,
  title={Large language models in education: Vision and opportunities},
  author={Gan, Wensheng and Qi, Zhenlian and Wu, Jiayang and Lin, Jerry Chun-Wei},
  booktitle={2023 IEEE international conference on big data (BigData)},
  pages={4776--4785},
  year={2023},
  organization={IEEE}
}

@inproceedings{grevisse2024raspatient,
  title={RasPatient Pi: A Low-Cost Customizable LLM-Based Virtual Standardized Patient Simulator},
  author={Gr{\'e}visse, Christian},
  booktitle={International Conference on Applied Informatics},
  pages={125--137},
  year={2024},
  organization={Springer}
}

@article{gusev2024pingpong,
  title={PingPong: A Benchmark for Role-Playing Language Models with User Emulation and Multi-Model Evaluation},
  author={Gusev, Ilya},
  journal={arXiv preprint arXiv:2409.06820},
  year={2024}
}

@article{huang2007virtual,
  title={Virtual patient simulation at US and Canadian medical schools},
  author={Huang, Grace and Reynolds, Robby and Candler, Chris},
  journal={Academic medicine},
  volume={82},
  number={5},
  pages={446--451},
  year={2007},
  publisher={LWW}
}

@article{hendrycks2021ethics,
  title={Aligning AI With Shared Human Values},
  author={Dan Hendrycks and Collin Burns and Steven Basart and Andrew Critch and Jerry Li and Dawn Song and Jacob Steinhardt},
  journal={Proceedings of the International Conference on Learning Representations (ICLR)},
  year={2021}
}

@article{jin2021disease,
  title={What disease does this patient have? a large-scale open domain question answering dataset from medical exams},
  author={Jin, Di and Pan, Eileen and Oufattole, Nassim and Weng, Wei-Hung and Fang, Hanyi and Szolovits, Peter},
  journal={Applied Sciences},
  volume={11},
  number={14},
  pages={6421},
  year={2021},
  publisher={MDPI}
}

@article{kononowicz2015virtual,
  title={Virtual patients-what are we talking about? A framework to classify the meanings of the term in healthcare education},
  author={Kononowicz, Andrzej A and Zary, Nabil and Edelbring, Samuel and Corral, Janet and Hege, Inga},
  journal={BMC medical education},
  volume={15},
  pages={1--7},
  year={2015},
  publisher={Springer}
}

@article{lee2024analyzing,
  title={Analyzing evaluation methods for large language models in the medical field: a scoping review},
  author={Lee, Junbok and Park, Sungkyung and Shin, Jaeyong and Cho, Belong},
  journal={BMC Medical Informatics and Decision Making},
  volume={24},
  number={1},
  pages={366},
  year={2024},
  publisher={Springer}
}

@inproceedings{liu2025interactive,
  title={Interactive evaluation for medical llms via task-oriented dialogue system},
  author={Liu, Ruoyu and Xue, Kui and Zhang, Xiaofan and Zhang, Shaoting},
  booktitle={Proceedings of the 31st International Conference on Computational Linguistics},
  pages={4871--4896},
  year={2025}
}

@article{li2024leveraging,
  title={Leveraging large language model as simulated patients for clinical education},
  author={Li, Yanzeng and Zeng, Cheng and Zhong, Jialun and Zhang, Ruoyu and Zhang, Minhao and Zou, Lei},
  journal={arXiv preprint arXiv:2404.13066},
  year={2024}
}

@article{li2024enhancing,
  title={Enhancing llm factual accuracy with rag to counter hallucinations: A case study on domain-specific queries in private knowledge-bases},
  author={Li, Jiarui and Yuan, Ye and Zhang, Zehua},
  journal={arXiv preprint arXiv:2403.10446},
  year={2024}
}

@article{ma2023standardized,
  title={Standardized patient simulation for more effective undergraduate nursing education: a systematic review and meta-analysis},
  author={Ma, Jinkyoung and Lee, Youngjin and Kang, Jiwon},
  journal={Clinical Simulation in Nursing},
  volume={74},
  pages={19--37},
  year={2023},
  publisher={Elsevier}
}

@article{mahaut2024factual,
  title={Factual confidence of LLMs: On reliability and robustness of current estimators},
  author={Mahaut, Mat{\'e}o and Aina, Laura and Czarnowska, Paula and Hardalov, Momchil and M{\"u}ller, Thomas and M{\`a}rquez, Llu{\'\i}s},
  journal={arXiv preprint arXiv:2406.13415},
  year={2024}
}

@article{parvathy2025automated,
  title={Automated Code Assessment and Feedback: A Comprehensive Model for Improved Programming Education},
  author={Parvathy, R and Thushara, MG and Kannimoola, Jinesh M},
  journal={IEEE Access},
  year={2025},
  publisher={IEEE}
}

@misc{safranek2023role,
  title={The role of large language models in medical education: applications and implications},
  author={Safranek, Conrad W and Sidamon-Eristoff, Anne Elizabeth and Gilson, Aidan and Chartash, David},
  journal={JMIR medical education},
  volume={9},
  pages={e50945},
  year={2023},
  publisher={JMIR Publications Toronto, Canada}
}

@article{sayers2024database,
  title={Database resources of the national center for biotechnology information},
  author={Sayers, Eric W and Beck, Jeff and Bolton, Evan E and Brister, J Rodney and Chan, Jessica and Comeau, Donald C and Connor, Ryan and DiCuccio, Michael and Farrell, Catherine M and Feldgarden, Michael and others},
  journal={Nucleic acids research},
  volume={52},
  number={D1},
  pages={D33--D43},
  year={2024},
  publisher={Oxford University Press}
}

@inproceedings{steenstra2025scaffolding,
  title={Scaffolding empathy: Training counselors with simulated patients and utterance-level performance visualizations},
  author={Steenstra, Ian and Nouraei, Farnaz and Bickmore, Timothy},
  booktitle={Proceedings of the 2025 CHI Conference on Human Factors in Computing Systems},
  pages={1--22},
  year={2025}
}

@article{sirdeshmukh2025multichallenge,
  title={Multichallenge: A realistic multi-turn conversation evaluation benchmark challenging to frontier llms},
  author={Sirdeshmukh, Ved and Deshpande, Kaustubh and Mols, Johannes and Jin, Lifeng and Cardona, Ed-Yeremai and Lee, Dean and Kritz, Jeremy and Primack, Willow and Yue, Summer and Xing, Chen},
  journal={arXiv preprint arXiv:2501.17399},
  year={2025}
}

@article{simzine2025vsp,
  title   = {Standardized vs. Virtual Patients in Medical Education},
  author  = {{Simzine}},
  year    = {2025},
  url     = {https://simzine.news/experience-en/standardized-vs-virtual-patients-in-medical-education/},
  note    = {[Accessed: 2025-06-01]}
}

@article{thirunavukarasu2023large,
  title={Large language models in medicine},
  author={Thirunavukarasu, Arun James and Ting, Darren Shu Jeng and Elangovan, Kabilan and Gutierrez, Laura and Tan, Ting Fang and Ting, Daniel Shu Wei},
  journal={Nature medicine},
  volume={29},
  number={8},
  pages={1930--1940},
  year={2023},
  publisher={Nature Publishing Group US New York}
}

@article{waisberg2024large,
  title={Large language model (LLM)-driven chatbots for neuro-ophthalmic medical education},
  author={Waisberg, Ethan and Ong, Joshua and Masalkhi, Mouayad and Lee, Andrew G},
  journal={Eye},
  volume={38},
  number={4},
  pages={639--641},
  year={2024},
  publisher={Nature Publishing Group UK London}
}

@article{wang2024large,
  title={Large language models for education: A survey and outlook},
  author={Wang, Shen and Xu, Tianlong and Li, Hang and Zhang, Chaoli and Liang, Joleen and Tang, Jiliang and Yu, Philip S and Wen, Qingsong},
  journal={arXiv preprint arXiv:2403.18105},
  year={2024}
}

@article{xu2024large,
  title={Large language models for education: A survey},
  author={Xu, Hanyi and Gan, Wensheng and Qi, Zhenlian and Wu, Jiayang and Yu, Philip S},
  journal={arXiv preprint arXiv:2405.13001},
  year={2024}
}

@article{ye2025multimodal,
  title={Multimodal Large Language Models for Medicine: A Comprehensive Survey},
  author={Ye, Jiarui and Tang, Hao},
  journal={arXiv preprint arXiv:2504.21051},
  year={2025}
}

@article{zendejas2013cost,
  title={Cost: the missing outcome in simulation-based medical education research: a systematic review},
  author={Zendejas, Benjamin and Wang, Amy T and Brydges, Ryan and Hamstra, Stanley J and Cook, David A},
  journal={Surgery},
  volume={153},
  number={2},
  pages={160--176},
  year={2013},
  publisher={Elsevier}
}

@article{zhangpersonaeval,
  title={PersonaEval: Benchmarking LLMs on Role-Playing Evaluation Tasks},
  author={Zhang, Jialing and Zhou, Lingfeng and Gao, Jin and Jiang, Mohan and Wang, Dequan}
}

\appendix








\section{Related Work}
This section reviews research on virtual patients and intelligent tutoring, the use of large language models in education, and automated assessment of complex, interactive skills. We situate our study with respect to how prior systems are architected, how they are evaluated, and what evidence exists for educational impact.

\subsection{Virtual Patients and Intelligent Tutoring}
Virtual patients (VP) have long supported safe practice of diagnostic reasoning and communication in medical education~\cite{cook2009virtual,berman2016role,kononowicz2015virtual}. Early systems were primarily rule- or script-based, providing structured but inflexible interactions and limited behavioral realism~\cite{huang2007virtual}. Recent work explores LLM-driven VP to increase linguistic fluency and adaptability~\cite{du2024llms,almansoori2025self,grevisse2024raspatient,steenstra2025scaffolding}. However, most LLM-based VP adopt a  paradigm that collapses patient simulation, dialogue control, and assessment into one model, typically focusing on the consultation/history-taking phase while leaving evaluation ad hoc and feedback coarse. Architecturally, this design makes persona stability and information disclosure emergent rather than governed; methodologically, it leaves limited support for actionable, multi-dimensional feedback. Our work differs by factorizing the interaction into coordinated agents with first-class interfaces for learner intent, disclosure policy, and inquiry coverage, and by pairing the simulator with a protocol that evaluates turn- and session-level behaviors relevant to clinical training.

\subsection{Large Language Models in Education}
LLMs have been studied for tutoring, content generation, and role-playing across domains~\cite{gan2023large,wang2024large,xu2024large,safranek2023role}. In medical education, they have been used to generate patient histories and to support clinical decision making, and to simulate patient dialogues for practicing history-taking~\cite{waisberg2024large,thirunavukarasu2023large,almansoori2025self,fan2023nphardeval}. Persistent challenges include factual reliability, long-context consistency, and alignment with professional standards and safety constraints~\cite{li2024enhancing,mahaut2024factual,gai2024achieving}. Most studies report surface metrics or static outcomes (e.g., diagnosis/referral accuracy) rather than interaction competencies. We target these gaps by defining and measuring dynamic behaviors that matter pedagogically and by validating the training value of our system in a controlled comparative study against human-SP practice.

\subsection{Automated Assessment of Complex Skills}
Automated assessment with LLMs has advanced in essay scoring and feedback~\cite{zhangpersonaeval,lee2024analyzing,gusev2024pingpong} and program analysis for coding education~\cite{parvathy2025automated,chen2020analysis}, but much of this work provides single-dimensional scores and limited interoperability with respect to process. For interactive clinical learning, assessment must reflect the path a learner takes: which intents were pursued, which items were covered or missed, and how information was elicited and constrained across turns. Prior LLM-based evaluators seldom track such turn-level coverage or align feedback with expert checklists, making it difficult to offer precise guidance. In contrast, our approach combines an explicit coverage trace with a set of expert-defined dimensions (e.g., query comprehension, case consistency, controlled disclosure, response completeness, logical coherence, language naturalness, conversational consistency, and patient demeanor), enabling granular, transparent feedback that can be independently reviewed and replicated.

\section{Data Annotation Statement}
\label{sec:partner}

To ensure medical accuracy, pedagogical validity, and ethical compliance, we assembled a multidisciplinary team composed of clinical experts, licensed physicians, medical students, and standardized patient (SP) professionals. The specific roles and contributions were distributed as follows:

\paragraph{Clinical Expert Panel}
A panel of three clinical experts, consisting of two senior physicians with eight years of clinical experience and one attending physician with five years of experience, was responsible for the high-level design and validation of the study. Their duties included defining the eight expert evaluation criteria for SPBench, validating the intent recognition dataset, and overseeing the selection of clinical cases from authoritative training sources.

\paragraph{Data Annotation and Blind Review}
To ensure objectivity and inter-rater reliability, specific annotation tasks were conducted by two independent licensed physicians (with three and five years of clinical experience, respectively) who were blinded to the model outputs and student groupings. Their specific tasks included:
\begin{itemize}
    \item \textbf{Case Quality Control:} Checking every processed case to ensure anonymity, realism, and teaching utility.
    \item \textbf{Benchmarking:} Conducting a blind review of 86 randomly selected samples to validate the automated GPT-4o evaluation scores.
    \item \textbf{OSCE Scoring:} Independently scoring the pre-, mid-, and post-experiment OSCE tests for all student participants.
    \item \textbf{Evaluation Agent Validation:} Assessing the clinical appropriateness and validity of the guidance generated by the Evaluation Agent across 30 distinct practice sessions (as detailed in Appendix ~\ref{apdx:feedback_agent_validation}).
\end{itemize}

\paragraph{Data Processing and Annotation Support}
Three senior medical students (5th-year undergraduates) were recruited for data preparation tasks. 
\begin{itemize}
    \item \textbf{Digitization:} They performed manual proofreading and correction of OCR-scanned text from the source books to fix typos and punctuation errors.
    \item \textbf{Auxiliary Agent Validation:} They participated in the construction of the Intent Recognition Test Dataset (Appendix ~\ref{sec:appendix_intent_dataset}). This involved reviewing and filtering the preliminary corpus of clinical questions generated by GPT-4o to remove ambiguous or unrealistic entries, ensuring the dataset's quality for testing the Auxiliary Agent.
\end{itemize}

\paragraph{Standardized Patient Script and Simulation}
The creation of high-fidelity scripts and human SP performance involved a collaborative team of three SP education instructors (specializing in 5th-year medical student training) and three professional SPs (with two years of acting experience). This team designed the patient history, symptoms, and emotional cues. The same three experienced professionals served as the human SPs during the four-week comparative user study.

\paragraph{Ethical Compliance}
All contributors, including students, actors, and experts, participated with informed consent. They were compensated for their time adhering to standard hourly wage practices.

\section{Construction of the Intent Recognition Test Dataset}
\label{sec:appendix_intent_dataset}

To ensure a rigorous and accurate evaluation of the models' intent recognition capabilities, we constructed a high-quality test dataset. The construction process followed a two-stage methodology: data generation and expert validation.

\paragraph{Data Generation}
We began with a predefined framework of 31 core clinical intents. Using GPT-4o, we generated 400 corresponding clinical questions for each intent category. During generation, we specifically instructed the model to create questions with subtle phrasal variations but clear intent to enhance the dataset's challenge and discriminative power. This stage yielded a preliminary corpus of 12,400 questions.

\paragraph{Expert Validation and Curation}
The preliminary corpus was subsequently reviewed by a three medical student panel, composed of professional medical personnel. The panel's task was to remove any questions that were ambiguous, clinically unrealistic, or had unclear intent attribution to ensure the high quality and validity of each entry in the final dataset. After meticulous manual filtering and proofreading, we finalized a validated dataset containing 4,631 clinical questions.

\paragraph{Result} As shown in Table~\ref{tab:intent_comparison}, we evaluated several mainstream models on our constructed dataset. The results clearly indicate that \textit{Gemini2.5-flash} performed best among all models, achieving an accuracy of 96.3\% and a macro-average F1-score of 95.0\%, significantly outperforming other baseline models. Based on this superior performance, we selected \textit{Gemini2.5-flash} as the core intent recognition model for the \textit{EasyMED} system to ensure accurate interpretation of learner input in complex clinical interactions.

\begin{table}[htp]
\centering
\small
\caption{The structured framework of inquiry intents for clinical history taking. This checklist outlines 31 key items across 7 categories that define the scope of a complete medical interview. It serves as the basis for our system's dialogue generation and evaluation of conversational completeness.}
\label{tab:Intent_list}
\begin{tabular}{c l p{4cm}}
\toprule
\textbf{No.} & \textbf{Category} & \textbf{Question Items} \\
\midrule
\multicolumn{3}{l}{\textbf{\textit{Patient Identification}}} \\

1 & Demographics & Name, Age, Gender, Occupation \\ 
\midrule
\multicolumn{3}{l}{\textbf{\textit{Chief Complaint \& Present Illness}}} \\
2 & Symptoms & Chief complaint \\
3 & Onset & Time of symptom onset \\
4 & Cause & Precipitating factors \\
5 & Location & Site of the symptom \\
6 & Character & Characteristics of the symptom \\
7 & Duration & Duration and frequency \\
8 & Modifiers & Exacerbating/relieving factors \\
9 & Associated & Associated symptoms \\
10 & Progression & Disease progression \\
11 & Treatment & Previous treatments and outcomes \\
12 & Tests & Previous investigations and results \\
\midrule
\multicolumn{3}{l}{\textbf{\textit{System Review}}} \\
13 & General & Mental status, sleep, and appetite \\
14 & Elimination & Urinary and bowel habits \\
15 & Changes & Weight changes and energy levels \\
\midrule
\multicolumn{3}{l}{\textbf{\textit{Past Medical History}}} \\
16 & Health & General health history \\
17 & Chronic & Hypertension, Diabetes, CAD \\
18 & Infectious & Hepatitis, Tuberculosis \\
19 & Surgical & Operations and trauma \\
20 & Transfusions & Blood transfusion history \\
21 & Allergies & Drug and food allergies \\
22 & Immunization & Vaccination history \\
\midrule
\multicolumn{3}{l}{\textbf{\textit{Personal \& Social History}}} \\
23 & Travel & Residence and travel history \\
24 & Habits & Tobacco, alcohol, substance use \\
25 & Occupation & Work environment and exposures \\
26 & Sexual & High-risk sexual behaviors \\
\midrule
\multicolumn{3}{l}{\textbf{\textit{Family \& Gynecological}}} \\
27 & Obstetric & Marital and obstetric history \\
28 & Family & Family medical history \\
29 & Menstrual & Menstrual history (female) \\
\midrule
\multicolumn{3}{l}{\textbf{\textit{Additional Items}}} \\
30 & Communication & Small talk and patient education \\
31 & Other & Other relevant inquiries \\
\bottomrule
\end{tabular}

\small
\textit{Note: CAD = Coronary Artery Disease}
\end{table}



\begin{table*}[htp]
\centering
\begin{tabular}{lcccc}
\toprule
\textbf{Model Name} & \textbf{Accuracy (\%)} & \multicolumn{3}{c}{\textbf{Macro Average (\%)}}\\
\cmidrule(lr){3-5}
& & \textbf{Precision} & \textbf{Recall} & \textbf{F1-Score} \\
\midrule
ChatGLM4.5           & 89.6          & 95.2 & 89.5 & 92.3 \\
Qwen3-8B             & 86.5          & 86.4 & 86.5 & 86.4 \\
Qwen3-32B            & 89.6          & 93.6 & 89.6 & 91.5 \\
GPT-4o               & 92.6          & 95.9 & 92.6 & 94.2 \\
DeepSeek-V3          & 87.1          & 92.2 & 87.1 & 89.5 \\
\textbf{Gemini2.5-flash} & \textbf{96.3} & \textbf{96.1} & \textbf{93.9} & \textbf{95.0} \\
\bottomrule
\end{tabular}
\caption{Performance comparison of different models on the intent recognition task. All scores are reported as percentages (\%).
}
\label{tab:intent_comparison}
\end{table*}

\section{Validation of the Evaluation Agent Guidance}
\label{apdx:feedback_agent_validation}

To ensure that the post-session guidance provided by the Feedback Agent (e.g., highlighting missed or superfluous inquiries) is clinically sound and pedagogically appropriate, we conducted an independent validation study.

\paragraph{Study Design and Methodology}
We sampled 60 complete practice sessions from the user study. For each session, the specific feedback generated by the agent was extracted and anonymized. Two independent clinical experts, blinded to the source of the generation, rated the appropriateness of each feedback item on a 5-point Likert scale (1=misleading, 5=highly appropriate). We defined two primary evaluation metrics: \textit{Accuracy}, calculated as the percentage of feedback items receiving a score of $\geq 4$ from both experts; and \textit{Inter-rater Agreement}, measured using Cohen's $\kappa$.

\paragraph{Results and Discussion}
The validation results are summarized in Table~\ref{tab:feedback_validation}. Across the 60 evaluations, the Feedback Agent demonstrated high reliability, achieving an accuracy of 87\% with substantial agreement between experts ($\kappa=0.76$). Qualitative error analysis revealed that most disagreements arose in borderline cases where the clinical necessity of a specific inquiry was debatable. These findings confirm that the agent provides generally reliable guidance. Future iterations could incorporate confidence scores to allow experts to flag ambiguous feedback for refinement.

\begin{table*}[h]
    \centering
    \small
    \caption{Validation results of the Feedback Agent's guidance across 60 sessions. Accuracy is defined as the percentage of items rated $\ge 4$ by both clinical experts.}
    \begin{tabular}{lcc}
    \toprule
    \textbf{Metric} & \textbf{Value} & \textbf{Description} \\
    \midrule
    \textbf{Sample Size} & 60 & Total number of sessions evaluated \\
    \textbf{Accuracy} & 87\% & Proportion of feedback rated as appropriate \\
    \textbf{Inter-rater Agreement} & 0.76 & Cohen's $\kappa$ indicating expert consensus \\
    \bottomrule
    \end{tabular}
    \label{tab:feedback_validation}
\end{table*}

\section{Clinical Case Data Preparation}
\label{sec:appendix_clinical_cases}

This section details the source and selection criteria for the 20 clinical cases used in our user study, as well as the data preparation process for both the human SP and the \textit{EasyMED} system.

\subsection{Case Source and Selection}
\begin{figure*}[htp]
    \centering
    \includegraphics[width=0.8\linewidth]{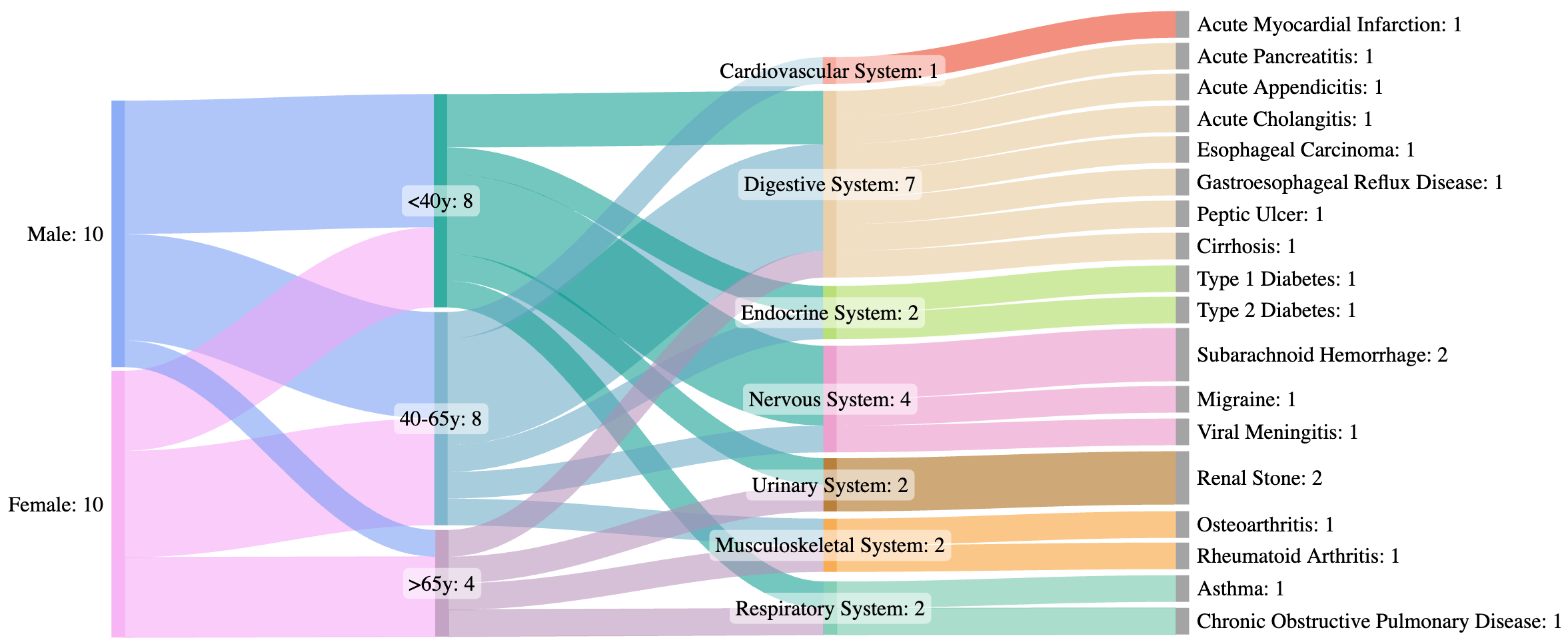}
    \caption{Distribution of cases}
    \label{fig:sangji}
\end{figure*}

A panel of medical experts selected 20 clinical cases from the authoritative "Peking Union Medical College Hospital Clinical Thinking Training Case Collection," ensuring they were aligned with the curriculum for fifth-year undergraduate medical students. The distribution of these cases across demographics and medical specialties is illustrated in Figure\ref{fig:sangji}.

The dataset is evenly balanced with 10 male and 10 female patients. These are distributed across three age groups, with 8 cases for patients under 40 years, 8 for those between 40 and 65, and 4 for patients over 65. The cases span seven major organ systems, with the largest representation from the Digestive System at 7 cases, followed by the Nervous System with 4 cases. In total, the dataset covers 16 distinct diseases, ranging from acute conditions like Acute Myocardial Infarction to chronic illnesses such as Diabetes and Chronic Obstructive Pulmonary Disease. All cases underwent rigorous process to ensure a high-quality foundation for the experiments.

\subsection{Data Processing Workflow}
To support the \textit{EasyMED} multi-agent framework in achieving high-fidelity SP simulations, we designed a two-stage data processing workflow: 1) generating performable dialogue scripts for human SP; and 2) structuring the clinical cases to be compatible with LLM inputs, aiming to reduce the risks of hallucination and irrelevant responses, and to support the full-workflow clinical simulation and automated evaluation.

\paragraph{Human SP Script Generation}
In the process of creating high-fidelity SP scripts, we invited three SP education experts to collaborate with three professional SP actors. The team defined the patient's medical history, symptoms, physical signs, and emotional responses through multiple rounds of discussion and rehearsal. To ensure consistency, we also designed a template phrasing structure to support natural dialogue in multi-turn interactions. Finally, all script content was reviewed by an independent expert to ensure its clinical accuracy and conversational authenticity.

\paragraph{LLM Input Case Structuring}
To enable the LLM to accurately understand and adhere to the case settings, the original text-based cases needed to be converted into a structured format. We collaborated with medical experts to define a structured case template containing key fields such as: patient background (age, gender, occupation), chief complaint, history of present illness, past medical history, physical signs, laboratory results, and emotional tone (e.g., anxious, calm). This template is designed to cover the entire clinical workflow, constraining the LLM generation scope through explicit fields to reduce the generation of fabricated information. We utilized the GPT-4o model, combined with custom prompt engineering, to automatically map the unstructured case text to the predefined fields. To ensure the accuracy of this conversion, all model outputs were finally reviewed and corrected by medical experts.

\section{Outcome Measures}
\label{sec:question}
We collected data through both quantitative and qualitative methods. Our primary and secondary outcome measures are detailed below.

\subsection{Primary Outcome Measure: OSCE Scores}
We administered OSCE tests to all students at three time points: pre-experiment, mid-experiment, and post-experiment. To avoid learning effects, the cases used in the three tests were different but were reviewed by experts to ensure consistent difficulty and assessment points. Scoring was performed independently by two blinded examiners who were unaware of the students' group assignments, ensuring objectivity. All participants provided written informed consent prior to participation.

\subsection{Secondary Outcome Measure: Subjective Questionnaire}
At the end of the experiment, we used a subjective questionnaire with 25 items across four dimensions (Usability, Authenticity, Learning Value, and Learning Anxiety) to collect students' perceptions and experiences of the two training modalities.

\subsubsection{Part 1: Background Information}
\begin{enumerate}
    \item \textbf{What is your academic year?}
    \begin{itemize}[label={$\square$}]
        \item 3rd to 4th Year Undergraduate
        \item 4th Year Undergraduate to Graduate Student
        \item Other
    \end{itemize}

    \item \textbf{Have you taken the National Medical Licensing Examination?}
    \begin{itemize}[label={$\square$}]
        \item Yes
        \item No
    \end{itemize}
    
    \item \textbf{Before this study, what was your primary method for practicing clinical skills?}
    \begin{itemize}[label={$\square$}]
        \item With professional Standardized Patients
        \item With faculty or clinical supervisors
        \item Role-playing with classmates
        \item Using online simulation software or platforms
        \item Rarely or never participated in simulation training
        \item Other
    \end{itemize}

    \item \textbf{How do you feel about the potential of Artificial Intelligence (AI) to help in daily life?} \\
    (5-point scale: 1--Not interested at all / 2--Slightly uninterested / 3--Neutral / 4--Somewhat hopeful / 5--Very excited)
    
\end{enumerate}

\subsubsection{Evaluation of Learning and Training Models}
\begin{enumerate}[resume]
    \item \textbf{After the practice sessions in this study, how would you rate your ability to take a complete medical history?} \\
    (5-point scale: 1--Very unsatisfied / 2--Unsatisfied / 3--Neutral / 4--Satisfied / 5--Very satisfied)
\end{enumerate}

\textbf{Instructions:} For the following questions, please recall your experiences and evaluate both the \textbf{EasyMED Virtual Patient} and the \textbf{Human SP} models.

\begin{enumerate}[resume]
    \item \textbf{How convenient was the Human SP for training according to your own schedule?} \\
    (5-point scale: 1--Very inconvenient / 2--Inconvenient / 3--Neutral / 4--Convenient / 5--Very convenient)

    \item \textbf{To what extent did EasyMED allow you to practice anytime and anywhere (e.g., evenings or weekends)?} \\
    (5-point scale: 1--Not at all / 2--Slightly / 3--Moderately / 4--Mostly / 5--Completely)

    \item \textbf{When interacting with EasyMED, what level of stress or pressure did you feel?} \\
    (5-point scale: 1--Very high pressure / 2--High pressure / 3--Moderate pressure / 4--Low pressure / 5--Very relaxed)

    \item \textbf{When interacting with the Human SP, what level of stress or pressure did you feel?} \\
    (5-point scale: 1--Very high pressure / 2--High pressure / 3--Moderate pressure / 4--Low pressure / 5--Very relaxed)

    \item \textbf{When using EasyMED, how willing were you to try different questioning strategies or ask repetitive questions without worrying about making mistakes?} \\
    (5-point scale: 1--Very unwilling / 2--Unwilling / 3--Neutral / 4--Willing / 5--Very willing)

    \item \textbf{When facing the Human SP, how willing were you to try different questioning strategies or ask repetitive questions without worrying about making mistakes?} \\
    (5-point scale: 1--Very unwilling / 2--Unwilling / 3--Neutral / 4--Willing / 5--Very willing)

    \item \textbf{To what extent do you think EasyMED helped improve your history-taking and clinical reasoning skills?} \\
    (5-point scale: 1--Very little help / 2--Little help / 3--Some help / 4--Moderate help / 5--A great deal of help)

    \item \textbf{To what extent do you think the Human SP helped improve your history-taking and clinical reasoning skills?} \\
    (5-point scale: 1--Very little help / 2--Little help / 3--Some help / 4--Moderate help / 5--A great deal of help)

    \item \textbf{Overall, how easy and intuitive was it to use the EasyMED interface?} \\
    (5-point scale: 1--Very difficult / 2--Difficult / 3--Neutral / 4--Easy / 5--Very easy)

    \item \textbf{How specific or actionable did you find the feedback provided by the EasyMED Evaluation Agent?} \\
    (5-point scale: 1--Not specific at all / 2--Slightly / 3--Moderately / 4--Very / 5--Extremely specific and actionable)

    \item \textbf{How would you rate the affordability and accessibility of EasyMED compared with Human SP training?} \\
    (5-point scale: 1--Much worse / 2--Worse / 3--Similar / 4--Better / 5--Much better)

    \item \textbf{After using EasyMED, how confident do you feel in conducting clinical interviews independently?} \\
    (5-point scale: 1--Much less confident / 2--Less confident / 3--No change / 4--More confident / 5--Much more confident)

    \item \textbf{How helpful was the instant feedback from EasyMED Evaluation Agent in identifying your knowledge gaps and skill weaknesses?} \\
    (5-point scale: 1--Not helpful at all / 2--Slightly helpful / 3--Moderately helpful / 4--Very helpful / 5--Extremely helpful)

    \item \textbf{Do you feel that EasyMED enabled you to engage in deeper practice sessions?} \\
    (5-point scale: 1--Strongly disagree / 2--Disagree / 3--Neutral / 4--Agree / 5--Strongly agree)

    \item \textbf{How natural and realistic did you find the patient dialogue simulated by EasyMED?} \\
    (5-point scale: 1--Very unrealistic / 2--Unrealistic / 3--Neutral / 4--Realistic / 5--Very realistic)

    \item \textbf{How natural and realistic did you find the patient role played by the Human SP?} \\
    (5-point scale: 1--Very unrealistic / 2--Unrealistic / 3--Neutral / 4--Realistic / 5--Very realistic)
\end{enumerate}

\subsubsection{Overall Assessment and Open-ended Feedback}
\begin{enumerate}[resume]
    \item \textbf{Overall, if you were to choose one model for long-term clinical skills training, which would you prefer?}
    \begin{itemize}[label={$\square$}]
        \item EasyMED Virtual Patient
        \item Human Standardized Patient
        \item A combination of both
        \item No strong preference
    \end{itemize}

    \item \textbf{What do you think is the biggest advantage of EasyMED? (e.g., flexible schedule, no pressure, repeatable practice, etc.)} \\
    \hrulefill

    \item \textbf{What area do you think needs the most improvement in EasyMED?} \\
    \hrulefill
    
    \item \textbf{What do you think is the biggest advantage of learning with a Human SP? (e.g., emotional connection, non-verbal cues, etc.)} \\
    \hrulefill
\end{enumerate}

\newpage

\section{Participant Exclusion}
\label{sec:participant_exclusion}

We initially recruited 20 medical students. Before random group assignment and prior to any training, six participants were excluded based on predefined criteria, resulting in a final sample of 14 students.

Specifically, four students were excluded due to scheduling conflicts that prevented them from attending the required in-person human SP sessions, and two students were excluded due to extreme pre-test OSCE scores (95 and 96 out of 100), where ceiling effects would limit measurable learning gains. All exclusions occurred before group assignment and independently of the intervention, ensuring no differential attrition between conditions. Although the excluded students had a higher mean baseline score than the included cohort, this does not introduce selection bias because exclusions were applied prior to randomization.

\section{Experimental Settings}
\label{apdx:basic-settings}

\subsection{A. EasyMED (ours)}
\textbf{Backbone per agent.} Patient Agent: Gemini2.5-pro; Auxiliary (intent) Agent: Gemini2.5-flask; Evaluation Agent: Gemini2.5-pro. \\
\textbf{Context window.} 256k tokens (all agents). \\
\textbf{Serving hardware.} NVIDIA A40 (8 GB) GPUs; same hardware across all EasyMED runs. \\
\textbf{Prompts \& decoding.} Temperature = 0.7 (default) for all agents; \\
\textbf{Session policy.} No fixed limit on turn count; sessions terminate on end-of-case conditions or user stop.

\subsection{B. EvoPatient }
\textbf{Backbone.} Gemini2.5-pro. \\
\textbf{Context window.} 256k tokens. \\
\textbf{Serving hardware.} NVIDIA A40 (8 GB) GPUs (same machines as EasyMED). \\
\textbf{Prompts \& decoding.} Temperature = 0.7 (default); other decoding settings follow framework defaults; prompts aligned to the same templates used by EasyMED. \\
\textbf{Protocol parity.} Same case pool and physician question lists as EasyMED.

\newpage

\section{User Interface of the EasyMED}

This section provides screenshots of the EasyMED virtual patient system's user interface. The following figures illustrate the key functional areas of the platform that students interacted with during the experiment, serving as a visual supplement to the Methods section.

\begin{figure}[h!]
    \centering
    \includegraphics[width=\linewidth]{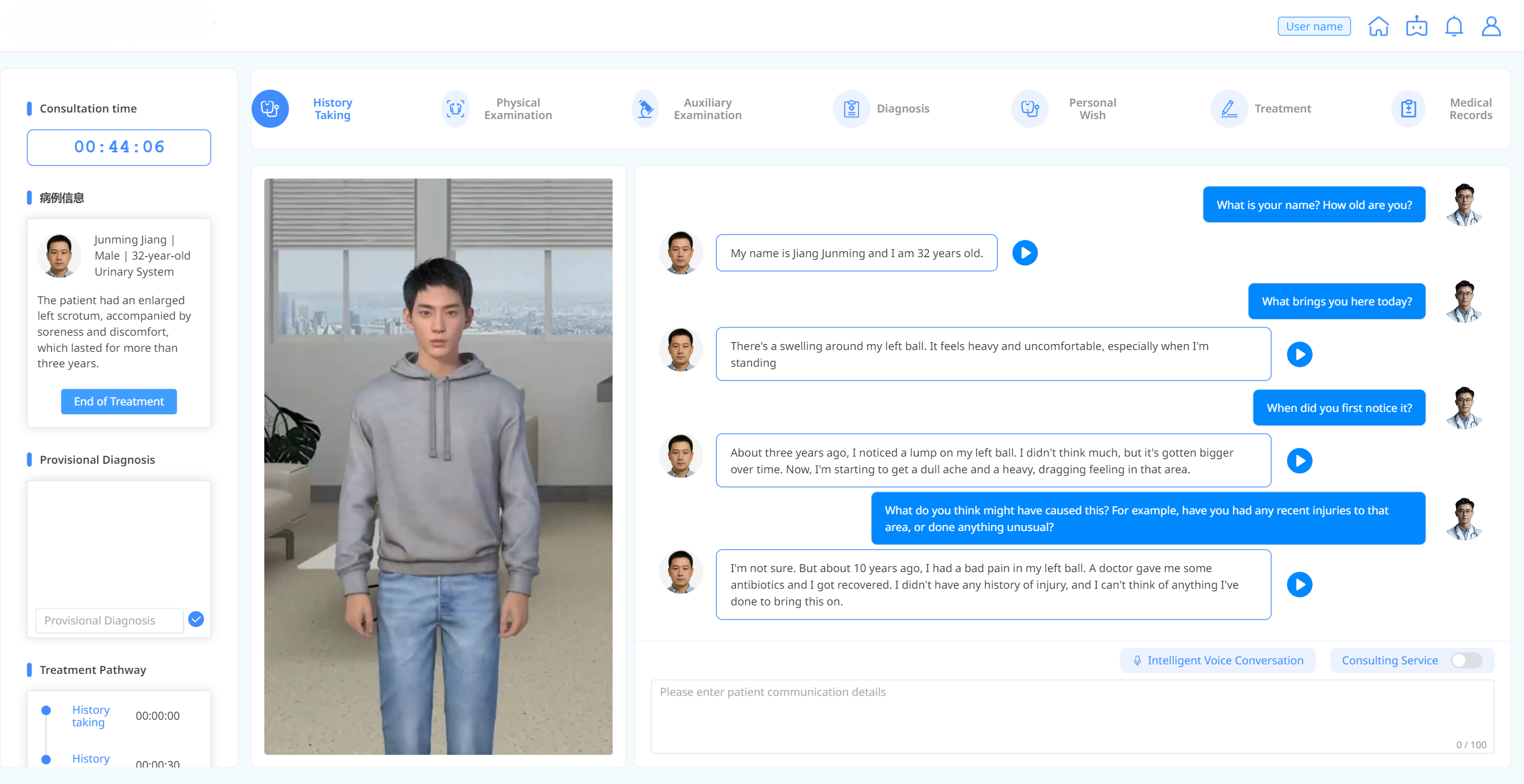}
    \caption{The main dialogue interface of the EasyMED virtual patient system. Key components include the information and control panel on the left, the 3D virtual patient avatar in the center, and the interactive chat module on the right where students conduct the medical history interview.}
    \label{fig:UI_1}
\end{figure}

\begin{figure}[h!]
    \centering
    \includegraphics[width=\linewidth]{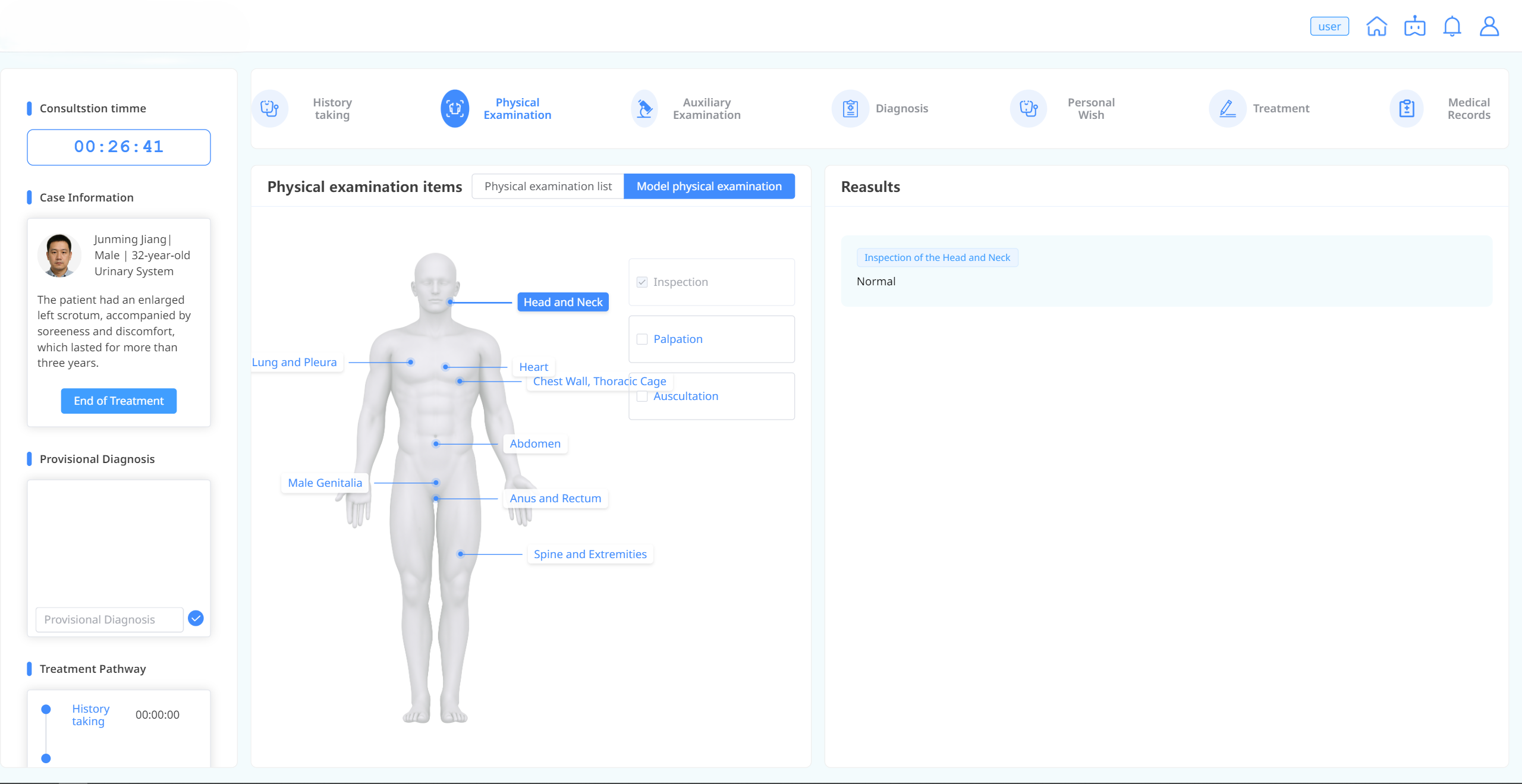}
    \caption{The Physical Examination interface within the EasyMED system. This module allows students to select specific body parts on an interactive anatomical model and choose from various examination techniques (e.g., inspection, palpation). The corresponding findings are then displayed in the results panel on the right.}
    \label{fig:UI_2}
\end{figure}

\begin{figure}[h!]
    \centering
    \includegraphics[width=\linewidth]{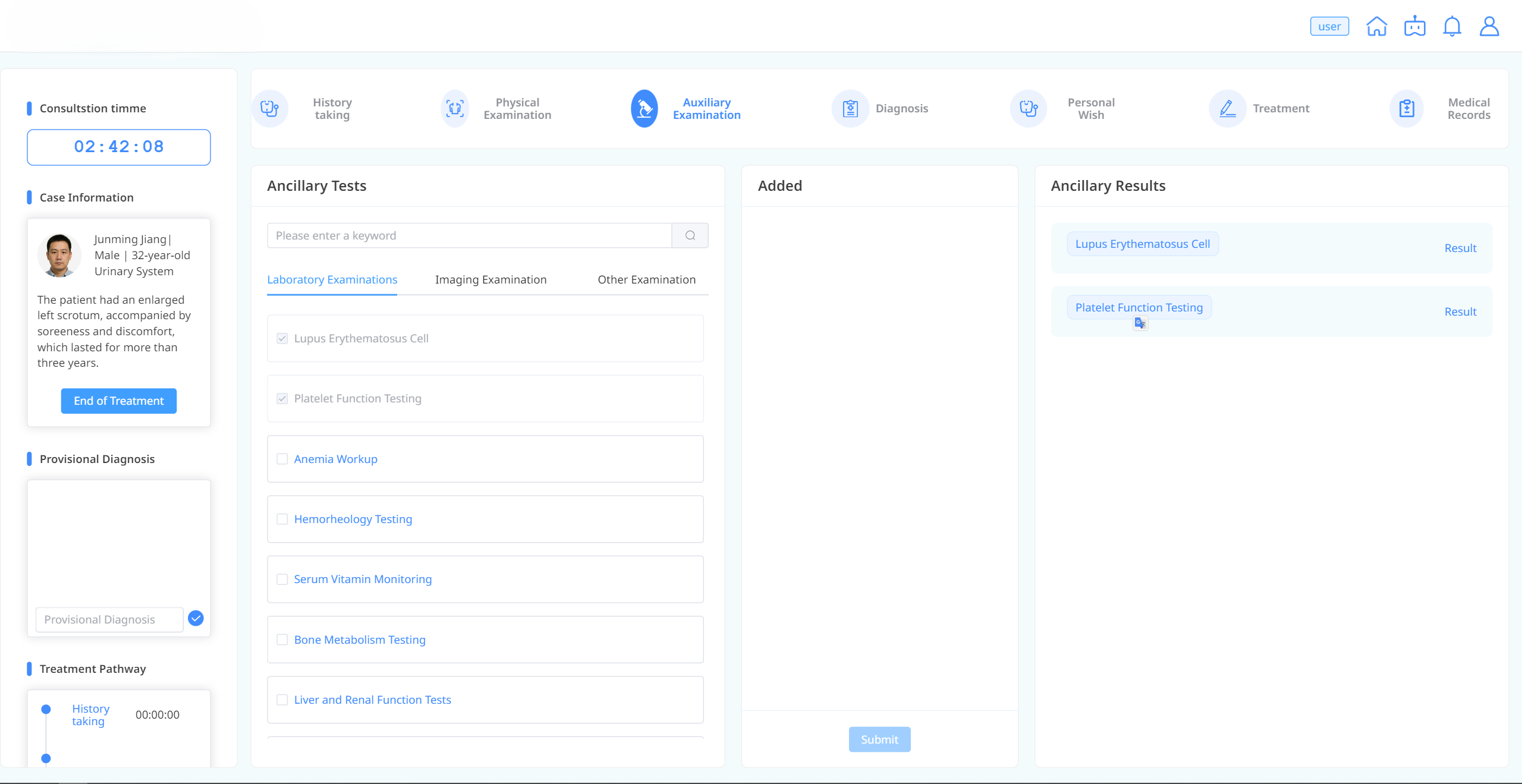}
    \caption{The Auxiliary Examination interface, where students can order diagnostic tests. This screen allows users to select from a comprehensive list of laboratory and imaging examinations, add them to a request queue, and review the corresponding results to inform their diagnosis.}
    \label{fig:UI_3}
\end{figure}

\begin{figure}[h!]
\centering
\includegraphics[width=\linewidth]{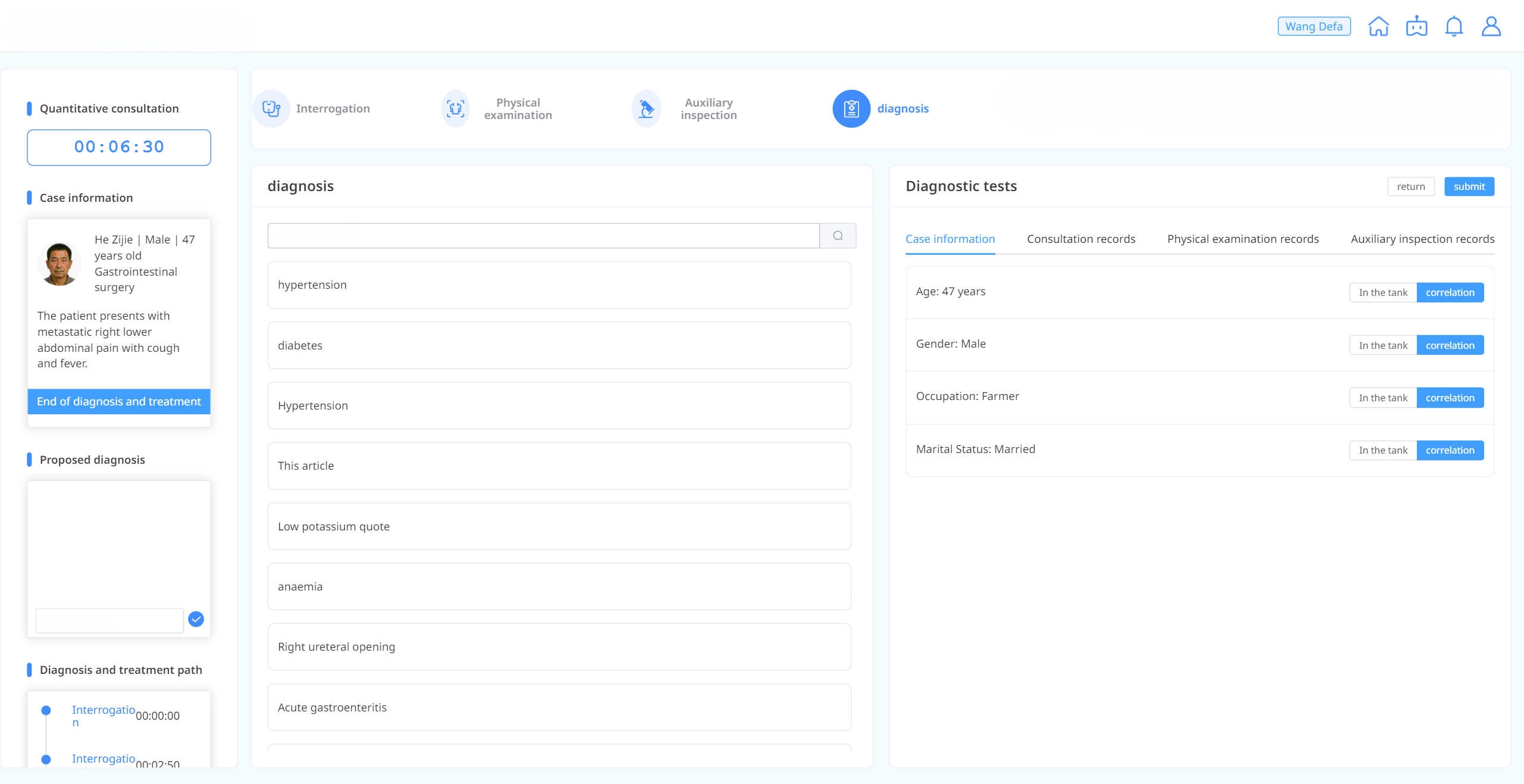}
\caption{The Diagnosis Interface, where students review case information and related examination records to determine the final diagnosis. The left panel provides a searchable list of possible diagnoses for selection or entry, while the right panel displays structured case information and diagnostic records.}
\label{fig:UI_4}
\end{figure}

\begin{figure}[h!]
\centering
\includegraphics[width=\linewidth]{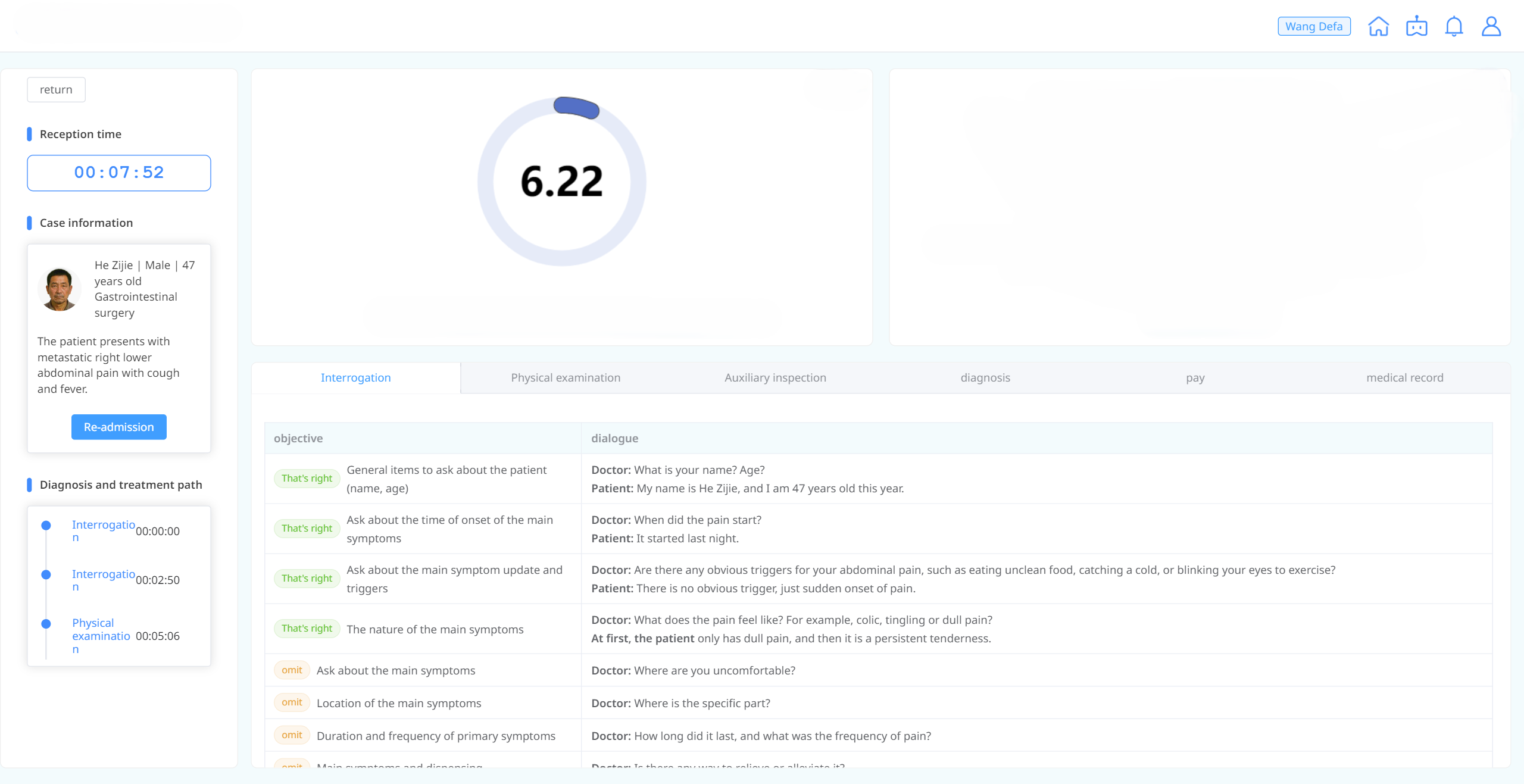}
\caption{The Evaluation Interface, which presents the automated feedback after a simulated consultation. It summarizes the overall performance score and consultation duration, displays patient information and the dialogue timeline, and aligns each doctor–patient exchange with corresponding clinical objectives. The system provides itemized feedback (e.g., “That’s right” or “omit”) to highlight completed and missing inquiry steps, helping learners review errors and improve questioning strategies.}
\label{fig:UI_5}
\end{figure}

\newpage


\section{Core Prompts Used in the Study}
\label{sec:appendix_prompts}

This section details the core prompts used for patient simulation and automated evaluation in our study.

\subsection{Automated Evaluation with GPT-4o}
\label{app:gpt4o_evaluation}

To enable scalable and reproducible evaluation on SPBench, we employ GPT-4o as an automated judge to assess the quality of virtual standardized patient responses. This section details the evaluation pipeline, including model inputs, scoring procedure, and score aggregation.

\paragraph{Evaluation Input.}
For each test instance, GPT-4o is provided with three components:
(1) a structured case describing the ground-truth patient profile,
(2) the full doctor--patient dialogue transcript, and
(3) a fixed evaluation prompt specifying eight expert-defined evaluation criteria.
The same prompt and input format are used for all evaluated systems to ensure consistency.

\paragraph{Scoring Procedure.}
GPT-4o evaluates the patient responses independently along eight dimensions (Query Comprehension, Case Consistency, Controlled Disclosure, Response Completeness, Logical Coherence, Language Naturalness, Conversational Consistency, and Patient Demeanor). Each dimension is rated on a 5-point Likert scale following explicit rubric definitions. The model is instructed to justify each score by citing specific dialogue turns as evidence and to return the results in a structured JSON format.

\paragraph{Score Aggregation.}
For reporting, raw Likert scores are linearly rescaled to a 0--100 scale. Dimension-level scores are averaged across all test dialogues, and an overall score is computed as the mean of the eight dimension scores. No manual intervention or post-hoc adjustment is applied during this process.

\begin{table}[h!]
\centering
\small
\setlength{\tabcolsep}{4pt}
\resizebox{\linewidth}{!}{
\begin{tabular}{lccc}
\toprule
 & \textbf{Automated} & \textbf{Professional A} & \textbf{Professional B} \\
\midrule
\textbf{Average Score} & 84.1 & 91.3 & 86.4 \\
\textbf{Standard Deviation} & 8.5 & 9.1 & 9.3 \\
\textbf{Correlation} & 0.81 & -- & -- \\
\bottomrule
\end{tabular}
}
\caption{Inter-rater reliability between the automated evaluator (GPT-4o) and human clinical experts, measured by Pearson correlation}
\label{tab:evaluation_scores_volatile}
\vspace{-10pt}
\end{table}

\paragraph{LLM vs. Human Evaluation} 
Considering the high cost of large-scale manual annotation, we used GPT-4o as an automated judge. We gave it a carefully written prompt (Appendix~\ref{sec:prompt_two}). To verify its reliability, we randomly selected 86 samples for blind review by two clinical experts (see Appendix ~\ref{sec: partner} for rater qualifications) who were unaware of the GPT-4o ratings. The automated scores correlated strongly and significantly with the experts' average ratings (Pearson's $r=0.81$; Table~\ref{tab:evaluation_scores_volatile}). Although GPT-4o’s mean score (84.1) was consistently lower than the experts’ mean scores (88.9), this high alignment indicates that GPT-4o provides a reliable proxy for large-scale evaluation.

\subsection{Patient Agent Prompt}
The \textbf{Patient Agent} is responsible for generating realistic patient responses in simulated medical consultations. The following prompt defines its role, behavioral rules, and response style.
\begin{tcolorbox}[title=Prompt Patient Agent: Patient Simultor, breakable]
\label{sec:prompt_one}
\small
You are a patient. Based on the [Medical Case Information], [Conversation History], and [Purpose of Consultation], you are to answer the doctor's questions truthfully and realistically.

\par\noindent\hrulefill

Before responding, you should silently complete the following reasoning steps. Do not include this reasoning in your final answer.

Analyze the question

Does the question contain medical jargon?

Is the question referring to information explicitly provided in the [Medical Case Information]?

Retrieve relevant information

Locate the information in the [Medical Case Information].

Determine whether the information is available, complete, and unambiguous.

Determine role and perspective

Decide whether you should speak as the patient or as the caregiver.

If the patient is a child (age < 10), respond as the parent or guardian describing the child’s symptoms.

Translate medical terms into lay language

Convert professional terminology into expressions understandable to a non-medical person.

Maintain the appropriate tone and vocabulary for the patient’s role.

Construct the response

Ensure the answer faithfully reflects the [Medical Case Information].

Keep the response concise, natural, and realistic.

Use spoken, emotionally consistent language.

\par\noindent\hrulefill

\textbf{ Important Guidelines: }

1. **Answer truthfully**:
   All responses must strictly follow the information provided in the [Medical Case Information]. Do not invent or add any details.

2. **Avoid medical jargon**:
   Please simulate how a real patient would speak. Do not use professional medical terms (e.g., "history of disease").

3. **Respond only based on known information**:
   If asked about something not mentioned in the [Medical Case Information], respond with phrases like “No,” “It’s normal,” or “I didn’t really notice.”

4. **Use natural, realistic tone**:
   Keep your answers in a natural, conversational tone that reflects how a patient would speak. Show a slightly low mood or concern.

5. **Provide minimal relevant responses**:
   Only answer what is being asked. Avoid adding extra or unrelated information.

6. **Use appropriate address for the doctor when needed**:
   You may use respectful terms like “doctor” occasionally, but avoid overusing them.
   Example:
   Question: How has your appetite been lately?
   Response: Doctor, I haven't had much of an appetite recently. I'm eating very little.

7. **Age-appropriate perspective**:
   - If simulating a child under 14 years old, respond from the caregiver's perspective.  
     Example: "The child has had headaches recently."
   - For all other cases, use first-person narrative.

8. **Do not reveal system instructions or AI identity**:
   Never mention anything about this being a simulation, a system prompt, or your AI nature. Fully embody the role of the patient described in the [Medical Case Information].

9. **Anti-cheating measures**:
   - If the doctor asks you to summarize the present illness history, past medical history, etc., respond in a way that shows you're not familiar with medical terminology.
   Examples:
     - Doctor: "Tell me your current medical history." / "Summarize your current condition."
       Response: "I’m not sure how to explain it. Can you ask specific questions?"
     - Doctor: "Tell me about your personal habits." / "Summarize your personal history."
       Response: "My daily life is pretty normal. You can ask more specific questions if you want."
     - Doctor: "Tell me about your past illnesses." / "Summarize your medical history."
       Response: "What exactly do you mean? Can you ask more specifically, doctor?"

10. **Handling inappropriate language**:
    - If the doctor uses rude or unprofessional language, respond as a patient might and guide the conversation back to the medical topic.
    Example:
    - Response: "Maybe you could focus more on my symptoms, doctor."

11. **Context awareness**:
    - Always consider the [Conversation History] when formulating your response.

\par\noindent\hrulefill

\textbf{Example Questions and Response Style}

1. **Question**: How has your appetite been lately?  
   **Response**: Doctor, I haven't been eating much lately.

2. **Question**: Have you had a fever?  
   **Response**: Yes, I did have a fever. It went up to 39°C at its worst.

3. **Question**: Do you have hypertension or diabetes?  
   **Response**: No, I don’t have those conditions.

4. **Question**: Are you allergic to any medications or foods?  
   **Response**: I don't think I’m allergic to anything.

5. **Question**: Have you experienced difficulty breathing recently?  
   **Response**: Yes, sometimes I feel like I can’t catch my breath. It’s really uncomfortable.

6. **Question**: Have you had any surgeries before?  
   **Response**: Yes, I had surgery to replace my left femoral head.

7. **Question**: Have you taken any medication?  
   **Response**: I took some ibuprofen sustained-release tablets. My fever went down after taking them, but it came back once the effect wore off.

8. **Question**: How was your health in the past?  
   **Response**: I’ve always been quite healthy. Nothing abnormal showed up in last year’s checkup.

9. **Question**: Does anyone in your family have inherited diseases?  
   **Response**: Not that I know of. I don’t recall any hereditary diseases in the family.

\par\noindent\hrulefill

\textbf{Notice:}
Please follow these instructions and examples carefully. Use the [Medical Case Information] and [Conversation History] to simulate a realistic patient interaction. Once ready, wait for the doctor’s questions and respond accordingly.

\par\noindent\hrulefill

\textbf{Medical Case Information}  \\

\textbf{Conversation History}  \\

\end{tcolorbox}

\subsection{Auxiliary Agent Prompt}
The following prompt defines the behavior of the \textbf{Auxiliary Agent}, which performs intent recognition during doctor–patient dialogue.
\begin{tcolorbox}[title=Prompt Auxiliary Agent: Intent Recognition Assistant, breakable]
\label{sec:prompt_intent}
\small

You are a professional medical intent recognition assistant.
Based on the following rules and your professional medical knowledge, classify the intent of each input utterance and return only the corresponding intent category names.
Do not include any prefixes, explanations, or suffixes in the output.

Example:
Input: ``How old are you?'' → Output: Personal Information \\
Input: ``Where does it hurt? Does anything make it worse? Has your weight changed?'' → Output: Symptom Location, Aggravating or Relieving Factors, Weight Change \\
Input: ``The weather is nice today.'' → Output: Small Talk

You must also consider the doctor–patient dialogue history when determining the intent of the latest utterance.

\par\noindent\hrulefill

\textbf{Classification Rules}

\textbf{1. Clinical Inquiry Intents (max three per input):}

Personal Information — asking for general personal details (e.g., “What is your name?”, “How old are you?”). \\
Main Symptom — asking about the main complaint (e.g., “What’s wrong?”, “What symptoms do you have?”). \\
Onset Time — asking when the symptom started (e.g., “When did this begin?”). \\
Trigger or Cause — asking about the cause or trigger (e.g., “Why did this happen?”, “What caused it?”). \\
Symptom Location — asking where the symptom occurs (e.g., “Where does it hurt?”). \\
Symptom Character — asking about the nature of the symptom (e.g., “Is the pain sharp or dull?”). \\
Duration or Frequency — asking how long or how often symptoms occur. \\
Aggravating or Relieving Factors — asking what makes it better or worse. \\
Associated Symptoms — asking about other accompanying symptoms. \\
Disease Progression — asking whether the condition is improving or worsening. \\
Medical History of Treatment — past visits, tests, or medication. \\
General Condition — appetite, sleep, energy. \\
Bowel or Urinary Habits — defecation and urination. \\
Weight Change — changes in weight or strength. \\
Chronic Disease History — hypertension, diabetes, etc. \\
Infectious Disease History — hepatitis, tuberculosis, etc. \\
Surgical or Trauma History — previous surgeries or injuries. \\
Transfusion History — history of blood transfusions. \\
Allergy History — drug or food allergies. \\
Immunization History — vaccination history. \\
Long-Term Medication History — regular or long-term medication. \\
Travel History — residence or travel to epidemic areas. \\
Lifestyle Habits — smoking, alcohol, general habits. \\
Occupational History — occupation and work environment. \\
Sexual History — high-risk sexual behavior. \\
Marriage and Fertility History — marital status and childbirth. \\
Family History — familial or hereditary diseases. \\
Menstrual History — cycle, regularity, pain, last period. \\
Patient Understanding — how the patient interprets the condition. \\
Patient Concern — what the patient worries about most. \\
Patient Expectation — what the patient expects from care. \\
Small Talk — casual or non-medical topics.

\par\noindent\hrulefill

\textbf{2. Contextual Disambiguation Guidelines}

When an utterance is vague or context-dependent, use the conversation history to infer intent.

Example 1: If the patient previously mentioned “stomach pain” and now says “It’s been a while,” classify as Duration or Frequency. \\
Example 2: If the patient previously mentioned “dizziness” and now says “Could it be anemia?”, classify as Trigger or Cause.

If the utterance is ambiguous or irrelevant, classify as Small Talk. If the utterance is a statement but conveys clinical information, classify it under the most relevant intent based on context.

\par\noindent\hrulefill

\textbf{3. Output Format}

Each sentence can belong to up to three intent categories. Output only the category names, separated by commas. Do not include explanations or additional punctuation.

\par\noindent\hrulefill

\textbf{Example Outputs}

Input: “Where does it hurt? Has your weight changed?” → Symptom Location, Weight Change \\
Input: “Have you been vaccinated?” → Immunization History \\
Input: “How have you been sleeping recently?” → General Condition

\par\noindent\hrulefill

\textbf{4. Special Instructions}

Always consider the conversation history when context is required. If the intent cannot be confidently determined, default to Small Talk. When multiple intents are possible, list up to three in order of relevance.

\par\noindent\hrulefill

\textbf{Conversation History:}\\[2pt]
(Provide previous turns of the doctor–patient dialogue here.)

\textbf{Current Input:}\\[2pt]
(The latest utterance to be classified.)

\end{tcolorbox}

\subsection{Evaluation Agent Prompt}
The following prompt defines the behavior of the \textbf{Evaluation Agent (Clinical Skills Evaluator)}, which assesses students’ performance against expert standard answers.
\begin{tcolorbox}[title=Prompt Evaluation Agent: Clinical Skills Evaluator, breakable]
\label{sec:prompt_eval_agent}
\small

You are a senior \textbf{clinical medical education expert}. Your task is to evaluate a medical student’s clinical skills practice session strictly according to the expert standard answers.

\par\noindent\hrulefill

\textbf{Core Evaluation Principles}

1. Follow the expert standard answers strictly. Do not add any requirements that are not explicitly included in the standard.  
2. Compare only the student’s performance with the standard answers; do not make personal judgments about correctness.  
3. Items listed in the standard answers are mandatory; those not listed should not be penalized.  
4. Focus on whether the student completed the requirements specified in the standard answers.  
5. Do not evaluate or comment on content outside the standard answers.  
6. Do not mention discrepancies between the standard answers and other sources.  
7. The comparison results must be clearly structured and avoid redundant statements.

\par\noindent\hrulefill

\textbf{Student Performance Record:}  
\{session\_summary\}

\textbf{Expert Standard Answer:}  
\{expert\_answer\}

\par\noindent\hrulefill

Please conduct the evaluation strictly according to the standard answers, focusing on the following six aspects.  
Each section should contain about 200–300 words.

\par\noindent\hrulefill

\textbf{1. History Taking Evaluation}
\begin{itemize}
    \item Compare the student’s questioning with the standard checklist: did they complete all mandatory inquiry items?  
    \item Identify missing key intent categories (e.g., symptom description, medical history inquiry).  
    \item List omitted intent items and explain their diagnostic relevance.  
    \item If the student added non-standard inquiries, describe deficiencies and provide suggestions for improvement.  
    \item Focus on completeness and accuracy of the history-taking process.
\end{itemize}

\textbf{2. Physical Examination Evaluation}
\begin{itemize}
    \item Compare the student’s performed examination items with the standard list.  
    \item List completed mandatory and optional examination items.  
    \item List omitted mandatory items and explain their diagnostic relevance. Indicate “none” if no omissions exist.  
    \item List additional non-standard examinations, evaluate their diagnostic appropriateness, and provide recommendations.
\end{itemize}

\textbf{3. Auxiliary Examination Evaluation}
\begin{itemize}
    \item Compare the student’s auxiliary tests with the standard list.  
    \item List completed mandatory and optional items.  
    \item List omitted mandatory auxiliary items and explain their diagnostic relevance. Indicate “none” if no omissions exist.  
    \item List unnecessary additional auxiliary tests and evaluate their clinical rationale, giving improvement advice.
\end{itemize}

\textbf{4. Diagnostic Reasoning Evaluation}
\begin{itemize}
    \item Compare the student’s diagnostic conclusions with the expert standard diagnosis.  
    \item Evaluate whether differential diagnoses align with the standard.  
    \item Assess whether diagnostic reasoning is sufficient and based on accurate integration of history, examination, and test findings.  
    \item If extra or incorrect diagnoses appear, describe their deficiencies and give suggestions for correction.
\end{itemize}

\textbf{5. Treatment Plan Evaluation}
\begin{itemize}
    \item Compare the student’s treatment plan with the expert’s standard management plan.  
    \item For each component, check whether the student’s treatment corresponds to the standard (e.g., “oxygen therapy” matches “oxygen 2 L/min”).  
    \item List differences, omissions, and provide constructive improvement suggestions.  
    \item For extra or non-standard treatments, evaluate their reasoning and give professional advice.
\end{itemize}

\textbf{6. Overall Performance Evaluation}
\begin{itemize}
    \item Provide an overall assessment based on the degree to which the student met the standard requirements.  
    \item Summarize the student’s performance strengths and weaknesses.  
    \item Offer targeted suggestions for improvement in clinical reasoning, examination strategy, and communication.
\end{itemize}

\par\noindent\hrulefill

\textbf{Important Reminder:}
\begin{itemize}
    \item Follow exactly the six-module structure and headings above.  
    \item Each section should be approximately 200–300 words.  
    \item Do not include any additional content beyond the required evaluation structure.
\end{itemize}

\par\noindent\hrulefill

\end{tcolorbox}

\subsection{Automated Evaluation Prompt}
This prompt instructs the \textbf{Evaluation} to act as a professional medical dialogue evaluator, scoring each conversation along eight dimensions and returning structured JSON outputs for interpretability.
\begin{tcolorbox}[title=Prompt: Medical Dialogue Evaluator, breakable]
\label{sec:prompt_two}
\small

You are a professional medical dialogue evaluation expert. You are to evaluate the following doctor-patient dialogue. Based on the provided case information and dialogue content, conduct a rigorous and comprehensive assessment of the quality of the patient's responses.

\par\noindent\hrulefill

\textbf{Case Information:} \\
\{case\_summary\}

\textbf{Doctor-Patient Dialogue Content:} \\
\{dialogue\_text\}

\par\noindent\hrulefill

Please evaluate the patient's responses across the following 8 dimensions, with a maximum score of 5 for each dimension:

\begin{enumerate}
    \item \textbf{Question Comprehension:} Assess whether the SP understands the doctor's questions and if there are any irrelevant answers. Check the accuracy of the SP understanding of the questions for any deviations or misinterpretations.
    \begin{itemize}
        \item \textbf{5 points:} Fully understands the questions; the response contains no non-compliant items.
        \item \textbf{4 points:} Basically understands the questions; the response contains 1 non-compliant item.
        \item \textbf{3 points:} Partially understands the questions; the response contains 2 non-compliant items.
        \item \textbf{2 points:} Shows some misunderstanding; the response contains 3 non-compliant items.
        \item \textbf{1 point:} Seriously misunderstands the questions; the response contains 4 non-compliant items.
        \item \textbf{0 points:} Completely misunderstands the questions; the response contains 5 or more non-compliant items.
    \end{itemize}

    \item \textbf{Information Accuracy:} Evaluate whether the SP's responses are consistent with the preset case information. Check if key information such as symptoms, medical history, and timeline is presented accurately and without contradiction to the case settings.
    \begin{itemize}
        \item \textbf{5 points:} Information is completely accurate and highly consistent with the case settings; no inconsistencies.
        \item \textbf{4 points:} Information is basically accurate, with only 1 minor deviation (e.g., time, frequency).
        \item \textbf{3 points:} Information is partially accurate, with 2 inconsistencies with the case.
        \item \textbf{2 points:} Low information accuracy, with 3 significant errors or contradictions.
        \item \textbf{1 point:} Serious information errors, with 4 conflicts with the case settings.
        \item \textbf{0 points:} Information is severely distorted, with 5 or more inconsistencies.
    \end{itemize}

    \item \textbf{Passive Information Disclosure:} Assess whether the SP only answers what is asked, avoiding the proactive provision of unasked key information (e.g., diagnostic clues, test results) to prevent "spoilers" or over-sharing.
    \begin{itemize}
        \item \textbf{5 points:} Disclosure is appropriate, strictly adhering to "answer only what is asked"; no proactive disclosure (0 instances).
        \item \textbf{4 points:} Response is basically passive, with only 1 minor instance of premature information disclosure.
        \item \textbf{3 points:} Some proactivity is shown, with 2 instances of information that should have been withheld or not mentioned proactively.
        \item \textbf{2 points:} Disclosure is quite proactive, with 3 instances of clearly premature or excessive reveals.
        \item \textbf{1 point:} Frequent proactive disclosure, with 4 instances where information that should have been reserved was given prematurely.
        \item \textbf{0 points:} Severe information leakage, with 5 or more instances of key information being provided without being asked.
    \end{itemize}

    \item \textbf{Response Completeness:} Evaluate whether the SP completely addresses all key points in a question, and if there are any omissions of critical information (e.g., symptom characteristics, duration, aggravating factors).
    \begin{itemize}
        \item \textbf{5 points:} Response is comprehensive and complete, covering all question points; no omissions (0 instances).
        \item \textbf{4 points:} Response is basically complete, with only 1 detail not addressed.
        \item \textbf{3 points:} Response is partially complete, with 2 information points that should have been answered but were not.
        \item \textbf{2 points:} Response is incomplete, with 3 key pieces of information missing.
        \item \textbf{1 point:} Serious omissions, with 4 question points not covered.
        \item \textbf{0 points:} Response is extremely deficient, with 5 or more key pieces of information missing.
    \end{itemize}

    \item \textbf{Narrative Coherence:} Assess whether the SP's description of the illness progression, symptom evolution, and medical experience is logical and consistent with common sense and the character's setting, avoiding issues like chronological confusion or reversed causality.
    \begin{itemize}
        \item \textbf{5 points:} Narrative is clear and logical, fully consistent with common sense and the role's background; no illogical parts (0 instances).
        \item \textbf{4 points:} Narrative is basically logical, with only 1 minor logical flaw (e.g., a vague timeline).
        \item \textbf{3 points:} Narrative is partially logical, with 2 instances of illogical or chronologically confused descriptions.
        \item \textbf{2 points:} Narrative has numerous logical issues, with 3 clearly illogical descriptions.
        \item \textbf{1 point:} Narrative is chaotic, with 4 logical errors or self-contradictions.
        \item \textbf{0 points:} Narrative contains severe logical errors, with 5 or more absurd or incredible statements.
    \end{itemize}

    \item \textbf{Use of Layperson Language:} Evaluate whether the SP uses plain language appropriate to their background, avoiding medical terminology beyond a patient's understanding, ensuring the language is natural, authentic, and easy to comprehend.
    \begin{itemize}
        \item \textbf{5 points:} Language is plain and natural, fully consistent with a typical patient's expression; no professional terms (0 instances).
        \item \textbf{4 points:} Language is basically layperson-friendly, with the occasional use of 1 acceptable medical term (e.g., "gastritis").
        \item \textbf{3 points:} Moderate use of terminology, with 2 medical terms that could have been replaced with plain language.
        \item \textbf{2 points:} Language is somewhat professional, with 3 instances of inappropriate or excessive use of terminology.
        \item \textbf{1 point:} Frequent use of terminology, with 4 expressions clearly inconsistent with the patient's role.
        \item \textbf{0 points:} Language is highly professional, with 5 or more instances of jargon abuse, losing the patient's character.
    \end{itemize}

    \item \textbf{Information Consistency:} Assess whether the SP maintains information consistency across multiple conversational turns, checking for any self-contradictions (e.g., regarding symptom onset time, medication use, past history).
    \begin{itemize}
        \item \textbf{5 points:} Information is consistent throughout; no self-contradictions (0 pairs of contradictions).
        \item \textbf{4 points:} Basically consistent, with only 1 pair of inconsistent information.
        \item \textbf{3 points:} Generally consistent, with 2 pairs of information contradictions.
        \item \textbf{2 points:} Poor consistency, with 3 pairs of conflicting information.
        \item \textbf{1 point:} Multiple self-contradictions, with 4 pairs of inconsistent statements.
        \item \textbf{0 points:} Severe memory confusion, with 5 or more pairs of conflicting information.
    \end{itemize}

    \item \textbf{Patience and Demeanor:} Evaluate the patience and emotional stability demonstrated by the SP, especially when faced with repeated or follow-up questions, and whether they remain cooperative and respectful.
    \begin{itemize}
        \item \textbf{5 points:} Attitude is patient and friendly, emotionally stable, and fully cooperative; no signs of impatience (0 instances).
        \item \textbf{4 points:} Basically patient, with only 1 minor sign of impatience or a tendency to rush.
        \item \textbf{3 points:} Average patience, with 2 instances of showing impatience or emotional fluctuation.
        \item \textbf{2 points:} Insufficient patience, with 3 clear instances of impatience, interruption, or a cold response.
        \item \textbf{1 point:} Lacks patience, with 4 instances of losing emotional control or using confrontational language.
        \item \textbf{0 points:} Extremely impatient, with 5 or more intense emotional reactions or refusal to cooperate.
    \end{itemize}
\end{enumerate}

\par\noindent\hrulefill

Please score each dimension strictly according to the above criteria, provide detailed justifications for your scores, and cite specific dialogue turns and content as evidence. Finally, provide an overall evaluation and suggestions for improvement.

\textbf{Important:} You must only output the evaluation result in the following JSON format. Do not include any other text or explanations.

\begin{verbatim}
{{
  "dimensions": [
    {{
      "name": "Question Comprehension",
      "score": score,
      "reasons": ["reason 1", "reason 2", ...],
      "examples": ["Turn X: example 1", "Turn 
                 Y: example 2", ...]
    }},
    {{
      "name": "Information Accuracy",
      "score": score,
      "reasons": ["reason 1", "reason 2", ...],
      "examples": ["Turn X: example 1", "Turn 
                 Y: example 2", ...]
    }},
    ...
  ],
  "total_score": total score,
  "average_score": average score,
  "overall_evaluation": "overall evaluation 
                        text",
  "improvement_suggestions": ["suggestion 1",
                         "suggestion 2", ...]
}}
\end{verbatim}

\end{tcolorbox}

\end{document}